\documentclass{article}
\newcommand{\TypeOfDoc}{CO}
\usepackage{lipsum}
\usepackage{times}
\usepackage{graphicx}
\usepackage{natbib}
\usepackage{color}
\usepackage{algorithm}
\usepackage{algorithmic}
\usepackage{rotating} 
\usepackage{array}
\usepackage{booktabs}
\usepackage[]{hyperref}
\usepackage[scaled=.8]{beramono} 
\usepackage{amsmath}
\usepackage{amsthm}
\usepackage{amsfonts}
\usepackage{amssymb}
\usepackage{bm}
\usepackage[capitalize]{cleveref}
\usepackage{array}
\usepackage{xspace}
\usepackage{mathtools}
\usepackage{subcaption}
\usepackage{tikz}
\usepackage{adjustbox}

\newcommand{\iftr}[1]{\ifthenelse{\equal{\TypeOfDoc}{TR}}{\color{black}#1\color{black}\xspace}{}} 

\newcommand{\ifco}[1]{\ifthenelse{\equal{\TypeOfDoc}{CO}}{\color{black}#1\color{black}\xspace}{}} 

\usepackage[accepted]{icml2017}

\newcolumntype{R}[1]{>{\begin{turn}{90}\begin{minipage}{#1}\scriptsize}l%
			<{\end{minipage}\end{turn}}%
}

\icmltitlerunning{Deep Decentralized Multi-task Multi-Agent RL under Partial Observability}

\graphicspath{{./figs/}}

\newtheorem{definition}{Definition}

\makeatletter
\newenvironment{proof*}[1][\proofname]{\par
	\pushQED{\qed}%
	\normalfont \partopsep=\z@skip \topsep=\z@skip
	\trivlist
	\item[\hskip\labelsep
	\itshape
	#1\@addpunct{.}]\ignorespaces
}{%
	\popQED\endtrivlist\@endpefalse
}
\makeatother

\newcommand{\bi}{\begin{itemize}}
\newcommand{\ei}{\end{itemize}}
\newcommand{\be}{\begin{enumerate}}
\newcommand{\ee}{\end{enumerate}}

\begin{document}
	
\twocolumn[
\icmltitle{Deep Decentralized Multi-task Multi-Agent Reinforcement Learning\\under Partial Observability}
\icmlsetsymbol{equal}{*}

\begin{icmlauthorlist}
	\icmlauthor{Shayegan Omidshafiei}{mit}
	\icmlauthor{Jason Pazis}{mit}
	\icmlauthor{Christopher Amato}{neu}
	\icmlauthor{Jonathan P. How}{mit}
	\icmlauthor{John Vian}{boeing}
\end{icmlauthorlist}

\icmlaffiliation{mit}{Laboratory for Information and Decision Systems (LIDS), MIT, Cambridge, MA, USA}
\icmlaffiliation{neu}{College of Computer and Information Science (CCIS), Northeastern University, Boston, MA, USA}
\icmlaffiliation{boeing}{Boeing Research \& Technology, Seattle, WA, USA}

\icmlcorrespondingauthor{Shayegan Omidshafiei}{shayegan@mit.edu}
\icmlkeywords{multi-agent multi-tasking, multi-agent reinforcement learning, decentralized deep recurrent Q-networks, ICML}

\vskip 0.3in
]

\printAffiliationsAndNotice{}  

\begin{abstract} 
	Many real-world tasks involve multiple agents with partial observability and limited communication. Learning is challenging in these settings due to local viewpoints of agents, which perceive the world as non-stationary due to concurrently-exploring teammates. Approaches that learn specialized policies for individual tasks face problems when applied to the real world: not only do agents have to learn and store distinct policies for each task, but in practice identities of tasks are often non-observable, making these approaches inapplicable. This paper formalizes and addresses the problem of multi-task multi-agent reinforcement learning under partial observability. We introduce a decentralized single-task learning approach that is robust to concurrent interactions of teammates, and present an approach for distilling single-task policies into a unified policy that performs well across multiple related tasks, without explicit provision of task identity.
\end{abstract} 
\section{Introduction}
In multi-task reinforcement learning (MTRL) agents are presented several related \emph{target} tasks \cite{taylor2009transfer, caruana1998multitask} with shared characteristics. Rather than specialize on a single task, the objective is to generalize performance across all tasks. For example, a team of autonomous underwater vehicles (AUVs) learning to detect and repair faults in deep-sea equipment must be able to do so in many settings (varying water currents, lighting, etc.), not just under the circumstances observed during training.

Many real-world problems involve multiple agents with partial observability and limited communication (e.g., the AUV example) \cite{DecPOMDPBook16}, but generating accurate models for these domains is difficult due to complex interactions between agents and the environment. Learning is difficult in these settings due to partial observability and local viewpoints of agents, which perceive the environment as non-stationary due to teammates' actions. Efficient learners must extract knowledge from past tasks to accelerate learning and improve generalization to new tasks. Learning specialized policies for individual tasks can be problematic, as not only do agents have to store a distinct policy for each task, but in practice face scenarios where the identity of the task is often non-observable.

Existing MTRL methods focus on single-agent and/or fully observable settings \cite{taylor2009transfer}. By contrast, this work considers cooperative, independent learners operating in partially-observable, stochastic environments, receiving feedback in the form of local noisy observations and joint rewards. This setting is general and realistic for many multi-agent domains. We introduce the multi-task multi-agent reinforcement learning (MT-MARL) under partial observability problem, where the goal is to maximize execution-time performance on a set of related tasks, without explicit knowledge of the task identity. Each MT-MARL task is formalized as a Decentralized Partially Observable Markov Decision Process (Dec-POMDP) \cite{bernstein2002complexity}, a general formulation for cooperative decision-making under uncertainty. MT-MARL poses significant challenges, as each agent must learn to coordinate with teammates to achieve good performance, ensure policy generalization across \emph{all} tasks, and conduct (implicit) execution-time inference of the underlying task ID to make sound decisions using local noisy observations. As typical in existing MTRL approaches, this work focuses on average asymptotic performance across all tasks \cite{caruana1998multitask, taylor2009transfer} and sample-efficient learning.

We propose a two-phase MT-MARL approach that first uses cautiously-optimistic learners in combination with Deep Recurrent Q-Networks (DRQNs) \mbox{\cite{hausknecht2015deep}} for action-value approximation. We introduce Concurrent Experience Replay Trajectories (CERTs), a decentralized extension of experience replay \cite{lin1992self,mnih2015human} targeting sample-efficient and stable MARL. This first contribution enables coordination in single-task MARL under partial observability. The second phase of our approach distills each agent's specialized action-value networks into a generalized recurrent multi-task network. Using CERTs and optimistic learners, well-performing distilled policies \cite{rusu2015policy} are learned for multi-agent domains. Both the single-task and multi-task phases of the algorithm are demonstrated to achieve good performance on a set of multi-agent target capture Dec-POMDP domains. The approach makes no assumptions about communication capabilities and is fully decentralized during learning and execution. To our knowledge, this is the first formalization of decentralized MT-MARL under partial observability. 

\section{Background}

\subsection{Reinforcement Learning}
Single-agent RL under full observability is typically formalized using Markov Decision Processes (MDPs) \cite{sutton1998reinforcement}, defined as tuple $\langle \mathcal{S}, \mathcal{A}, \mathcal{T}, \mathcal{R}, \gamma \rangle$. At timestep $t$, the agent with state $s \in \mathcal{S}$ executes action $a \in \mathcal{A}$ using policy $\pi(a|s)$, receives reward $r_{t} = \mathcal{R}(s) \in \mathbb{R}$, and transitions to state $s' \in \mathcal{S}$ with probability $P(s'|s,a) = \mathcal{T}(s,a,s')$. Denoting discounted return as $R_t = \sum_{t'=t}^{H}\gamma^{t'-t}r_{t}$, with horizon $H$ and discount factor $\gamma \in [0,1)$, the action-value (or Q-value) is defined as $Q^{\pi}(s,a) = \mathbb{E}_{\pi}[R_t|s_t=s,a_t=a]$. Optimal policy $\pi^{*}$ maximizes the Q-value function, $Q^{\pi^{*}}(s,a) = \max_{\pi}Q(s,a)$. In RL, the agent interacts with the environment to learn $\pi^{*}$ \emph{without} explicit provision of the MDP model. Model-based methods first learn $\mathcal{T}$ and $\mathcal{R}$, then use a planner to find $Q^{\pi^{*}}$. Model-free methods typically directly learn Q-values or policies, so can be more space and computation efficient.

Q-learning \cite{watkins1992q} iteratively estimates the optimal Q-value function using backups,
$Q(s,a) = Q(s,a) + \alpha[r + \gamma \max_{a'} Q(s',a')-Q(s,a)]$,
where $\alpha \in [0,1)$ is the learning rate and the term in brackets is the temporal-difference (TD) error. Convergence to $Q^{\pi^*}$ is guaranteed in the tabular (no approximation) case provided sufficient state/action space exploration; however, tabulated learning is unsuitable for problems with large state/action spaces. Practical TD methods instead use function approximators \cite{gordon1995stable} such as linear combinations of basis functions or neural networks, leveraging inductive bias to execute similar actions in similar states. Deep Q-learning is a state-of-the-art approach using a Deep Q-Network (DQN) for Q-value approximation \cite{mnih2015human}. At each iteration $j$, experience tuple $\langle s,a,r,s' \rangle$ is sampled from replay memory $\mathcal{M}$ and DQN parameters $\theta$ are updated to minimize loss $L_j(\theta_j) = \mathbb{E}_{(s,a,r,s')\sim \mathcal{M}} \lbrack(r+\gamma \max_{a'}Q(s',a';\hat{\theta}_{j})-Q(s,a;\theta_j))^{2}\rbrack$. Replay memory $\mathcal{M}$ is a first-in first-out queue containing the set of latest experience tuples from $\epsilon$-greedy policy execution. \emph{Target network} parameters $\hat{\theta}_{j}$ are updated less frequently and, in combination with experience replay, are critical for stable deep Q-learning.

Agents in partially-observable domains receive observations of the latent state. Such domains are formalized as Partially Observable Markov Decision Processes (POMDPs), defined as $\langle \mathcal{S}, \mathcal{A}, \mathcal{T}, \mathcal{R}, \Omega, \mathcal{O}, \gamma \rangle$ \cite{kaelbling1998planning}. After each transition, the agent observes $o \in \Omega$ with probability $P(o|s',a) = \mathcal{O}(o,s',a)$. Due to noisy observations, POMDP policies map observation histories to actions. As Recurrent Neural Networks (RNNs) inherently maintain an internal state $h_t$ to compress input history until timestep $t$, they have been demonstrated to be effective for learning POMDP policies \cite{wierstra2007solving}. Recent work has introduced Deep Recurrent Q-Networks (DRQNs) \cite{hausknecht2015deep}, combining Long Short-Term Memory (LSTM) cells \cite{hochreiter1997long} with DQNs for RL in POMDPs. Our work extends this single-task, single-agent approach to the multi-task, multi-agent setting.

\subsection{Multi-agent RL}
Multi-agent RL (MARL) involves a set of agents in a shared environment, which must learn to maximize their individual returns \cite{bucsoniu2010multi}. Our work focuses on cooperative settings, where agents share a joint return. \citet{claus1998dynamics} dichotomize MARL agents into two classes: Joint Action Learners (JALs) and Independent Learners (ILs). JALs observe actions taken by \emph{all} agents, whereas ILs only observe \emph{local} actions. As observability of joint actions is a strong assumption in partially observable domains, ILs are typically more practical, despite having to solve a more challenging problem \cite{claus1998dynamics}. Our approach utilizes ILs that conduct both learning \emph{and} execution in a decentralized manner.


Unique challenges arise in MARL due to agent interactions during learning \cite{bucsoniu2010multi,matignon2012independent}. Multi-agent domains are non-stationary from agents' local perspectives, due to teammates' interactions with the environment. ILs, in particular, are susceptible to \emph{shadowed equilibria}, where local observability and non-stationarity cause locally optimal actions to become a globally sub-optimal joint action \cite{fulda2007predicting}. Effective MARL requires each agent to tightly coordinate with fellow agents, while also being robust against destabilization of its own policy due to environmental non-stationarity. Another desired characteristic is robustness to \emph{alter-exploration}, or drastic changes in policies due to exploratory actions of teammates \cite{matignon2012independent}.



\subsection{Transfer and Multi-Task Learning}

Transfer Learning (TL) aims to generalize knowledge from a set of \emph{source tasks} to a \emph{target task} \cite{pan2010survey}. In single-agent, fully-observable RL, each task is formalized as a distinct MDP (i.e., MDPs and tasks are synonymous) \cite{taylor2009transfer}. While TL assumes sequential transfer, where source tasks have been previously learned and even may not be related to the target task, Multi-Task Reinforcement Learning (MTRL) aims to learn a policy that performs well on related target tasks from an underlying task distribution \cite{caruana1998multitask, pan2010survey}. MTRL tasks can be learned simultaneously or sequentially \cite{taylor2009transfer}. MTRL directs the agent's attention towards pertinent training signals learned on individual tasks, enabling a \emph{unified} policy to generalize well across all tasks. MTRL is most beneficial when target tasks share common features \cite{wilson2007multi}, and most challenging when the task ID is not explicitly specified to agents during execution -- the setting addressed in this paper.

\section{Related Work}



\subsection{Multi-agent RL}
\citet{bucsoniu2010multi} present a taxonomy of MARL approaches. Partially-observable MARL has received limited attention. Works include model-free gradient-ascent based methods \cite{peshkin2000learning,Dutech01}, simulator-supported methods to improve policies using a series of linear programs \cite{wu2012rollout}, and model-based approaches where agents learn in an interleaved fashion to reduce destabilization caused by concurrent learning \cite{banerjee2012sample}. 
Recent scalable methods use Expectation Maximization to learn finite state controller (FSC) policies \cite{Wu13,LiuIJCAI15,LiuAAAI16}. 




Our approach is most related to IL algorithms that learn Q-values, as their representational power is more conducive to transfer between tasks, in contrast to policy tables or FSCs. The  majority of existing IL approaches assume full observability. \citet{matignon2012independent} survey these approaches, the most straightforward being Decentralized Q-learning \cite{tan1993multi}, where each agent performs independent Q-learning. This simple approach has some empirical success \cite{matignon2012independent}.  Distributed Q-learning \cite{lauer2000algorithm} is an optimal algorithm for deterministic domains; it updates Q-values only when they are guaranteed to increase, and the policy only for actions that are no longer greedy with respect to Q-values.  \citet{bowling2002multiagent} conduct Policy Hill Climbing using the Win-or-Learn Fast heuristic to decrease (increase) each agent's learning rate when it performs well (poorly). Frequency Maximum Q-Value heuristics \cite{kapetanakis2002reinforcement} bias action selection towards those consistently achieving max rewards. Hysteretic Q-learning \cite{matignon2007hysteretic} addresses miscoordination using cautious optimism to stabilize policies while teammates explore. Its track record of empirical success against complex methods \cite{xu2012multiagent,matignon2012independent,barbalios2014robust} leads us to use it as a foundation for our MT-MARL approach. \citet{foerster2016learning} present architectures to learn communication protocols for Dec-POMDP RL, noting best performance using a centralized approach with inter-agent backpropagation and parameter sharing. They also evaluate a model combining Decentralized Q-learning with DRQNs, which they call Reinforced Inter-Agent Learning. Given the decentralized nature of this latter model (called Dec-DRQN herein for clarity), we evaluate our method against it. Concurrent to our work, \citet{foerster2017stabilising} investigated an alternate means of stabilizing experience replay for the centralized learning case.

\subsection{Transfer and Multi-task RL}
\citet{taylor2009transfer} and \citet{torrey2009transfer} provide excellent surveys of transfer and multi-task RL, which almost exclusively target single-agent, fully-observable settings. \citet{tanaka2003multitask} use first and second-order statistics to compute a prioritized sweeping metric for MTRL, enabling an agent to maximize lifetime reward over task sequences. \citet{fernandez2006probabilistic} introduce an MDP policy similarity metric, and learn a \emph{policy library} that generalizes well to tasks within a shared domain. \citet{wilson2007multi} consider TL for MDPs, learning a Dirichlet Process Mixture Model over source MDPs, used as an informative prior for a target MDP. They extend the work to multi-agent MDPs by learning characteristic agent roles \cite{wilson2008learning}. \citet{brunskill2013sample} introduce an MDP clustering approach that reduces negative transfer in MTRL, and prove reduction of sample complexity of exploration using transfer. \citet{taylor2013transfer} introduce \emph{parallel transfer} to accelerate multi-agent learning using inter-agent transfer.  
Recent work extends the notion of neural network distillation \cite{hinton2015distilling} to DQNs for single-agent, fully-observable MTRL, first learning a set of specialized teacher DQNs, then distilling teachers to a single multi-task network \cite{rusu2015policy}. The efficacy of the distillation technique for single-agent MDPs with large state spaces  leads our work to use it as a foundation for the proposed MT-MARL under partial observability approach.

\section{Multi-task Multi-agent RL}


This section introduces MT-MARL under partial observability. We formalize single-task MARL using the Decentralized Partially Observable Markov Decision Process (Dec-POMDP), defined as $\langle \mathcal{I}, \mathcal{S}, \bm{\mathcal{A}}, \mathcal{T}, \mathcal{R}, \bm{\Omega}, \mathcal{O}, \gamma \rangle$, where $\mathcal{I}$ is a set of $n$ agents, $\mathcal{S}$ is the state space, $\bm{\mathcal{A}} = \times_i \mathcal{A}^{(i)}$ is the joint action space, and $\bm{\Omega} = \times_i \Omega^{(i)}$ is the joint observation space \cite{bernstein2002complexity}.\footnote{Superscripts indicate \emph{local} parameters for agent $i \in \mathcal{I}$.} Each agent $i$ executes action $a^{(i)} \in \mathcal{A}^{(i)}$, where joint action $\bm{a} = \langle a^{(1)}, \ldots, a^{(n)} \rangle$ causes environment state $s \in \mathcal{S}$ to transition with probability $P(s'|s,\bm{a}) = \mathcal{T}(s,\bm{a},s')$. At each timestep, each agent receives observation $o^{(i)} \in \Omega^{(i)}$, with joint observation probability $P(\bm{o}|s',\bm{a}) = \mathcal{O}(\bm{o},s',\bm{a})$, where $\bm{o} = \langle o^{(1)}, \ldots, o^{(n)} \rangle$. Let local observation history at timestep $t$ be $\vec{o_t}^{(i)} = (o^{(i)}_{1},\ldots,o^{(i)}_{t})$, where $\vec{o_t}^{(i)} \in \vec{O_t}^{(i)}$. Single-agent policy $\pi^{(i)}: \vec{O_t}^{(i)} \mapsto A^{(i)} $ conducts action selection, and the joint policy is denoted $\bm{\pi} = \langle \pi^{(1)}, \ldots, \pi^{(n)} \rangle $. For simplicity, we consider only pure joint policies, as finite-horizon Dec-POMDPs have at least one pure joint optimal policy \cite{oliehoek2008optimal}. The team receives a joint reward $r_t = \mathcal{R}(s_t,\bm{a}_t) \in \mathbb{R}$ at each timestep $t$, the objective being to maximize the value (or expected return), $V = \mathbb{E}[\sum_{t=0}^{H}\gamma^{t}r_t]$. While Dec-POMDP \emph{planning} approaches assume agents do not observe intermediate rewards, we make the typical RL assumption that they do. This assumption is consistent with prior work in MARL \cite{banerjee2012sample,peshkin2000learning}.


ILs provide a scalable way to learn in Dec-POMDPs, as each agent's policy maps local observations to actions. However, the domain appears \emph{non-stationary} from the perspective of each Dec-POMDP agent, a property we formalize by extending the definition by \citet{laurent2011world}.
\begin{definition} 
	Let $\bm{a}^{-(i)} = \bm{a} \setminus \{a^{(i)}\}$. Local decision process for agent $i$ is stationary if, for all timesteps $t,u \in \mathbb{N}$, 
	\begin{equation}\label{eq:stationary_trans}
	\thinmuskip=0mu 
	\sum_{\mathclap{\hspace{13pt}\bm{a}_t^{-(i)}\in \bm{\mathcal{A}}^{-(i)}}} P(s'|s,\langle a^{(i)}, \bm{a}_{t}^{-(i)} \rangle ) 
	=  \sum_{\mathclap{\hspace{13pt}\bm{a}_u^{-(i)}\in \bm{\mathcal{A}}^{-(i)}}} P(s'|s, \langle a^{(i)}, \bm{a}_{u}^{-(i)} \rangle),
	\end{equation}
	and 	 
 	\begin{equation}\label{eq:stationary_obs}
		\thinmuskip=0mu 
		\sum_{\mathclap{\hspace{22pt}\bm{a}_t^{-(i)}\in \bm{\mathcal{A}}^{-(i)}}} P(o^{(i)}|s',\langle a^{(i)}, \bm{a}_{t}^{-(i)} \rangle) =  \sum_{\mathclap{\bm{a}_u^{-(i)}\in \bm{\mathcal{A}}^{-(i)}}} P(o^{(i)}|s',\langle a^{(i)}, \bm{a}_{u}^{-(i)} \rangle).
 	\end{equation}
\end{definition}



Letting $\bm{\pi}^{-(i)} = \bm{\pi} \setminus \{\pi^{(i)}\}$, non-stationarity from the local perspective of agent $i$ follows as in general $\bm{a}_t^{-(i)} = \bm{\pi}^{-(i)}(\vec{\bm{o}}_t) \neq \bm{\pi}^{-(i)}(\vec{\bm{o}}_u) = \bm{a}_u^{-(i)}$, which causes violation of \eqref{eq:stationary_trans} and \eqref{eq:stationary_obs}. Thus, MARL extensions of single-agent algorithms that assume stationary environments, such as Dec-DRQN, are inevitably ill-fated. This motivates our decision to first design a single-task, decentralized MARL approach targeting non-stationarity in Dec-POMDP learning.

The MT-MARL problem in partially observable settings is now introduced by extending the single-agent, fully-observable definition of \citet{fernandez2006probabilistic}.

\begin{definition}	A partially-observable MT-MARL Domain $\mathcal{D}$ is a tuple $\langle \mathcal{I}, \mathcal{S}, \bm{\mathcal{A}}, \bm{\Omega}, \gamma \rangle$, where $\mathcal{I}$ is the set of agents, $\mathcal{S}$ is the environment state space, $\bm{\mathcal{A}}$ is the joint action space, $\bm{\Omega}$ is the joint observation space, and $\gamma$ is the discount factor.
\end{definition}

\begin{definition} A partially-observable MT-MARL Task $T_j$ is a tuple $\langle \mathcal{D}, \mathcal{T}_j, \mathcal{R}_j, \mathcal{O}_j \rangle$, where $\mathcal{D}$ is a shared underlying domain; $\mathcal{T}_j$, $\mathcal{R}_j$, $\mathcal{O}_j$ are, respectively, the task-specific transition, reward, and observation functions.
\end{definition}

In MT-MARL, each episode \mbox{$e \in \{1,\ldots,E\}$} consists of a randomly sampled Task $T_j$ from domain $\mathcal{D}$. The team observes the task ID, $j$, during learning, but not during execution. The objective is to find a joint policy that maximizes average empirical execution-time return in all $E$ episodes, $\bar{V} = \frac{1}{E} \sum_{e=0}^{E} \sum_{t=0}^{H_e}\gamma^{t}\mathcal{R}_e(s_t,\bm{a}_t)$, where $H_e$ is the time horizon of episode $e$.

\section{Approach}
This section introduces a two-phase approach for partially-observable MT-MARL; the approach first conducts single-task specialization, and subsequently unifies task-specific DRQNs into a joint policy that performs well in all tasks.

\subsection{Phase I: Dec-POMDP Single-Task MARL}
As Dec-POMDP RL is notoriously complex (and solving for the optimal policy is NEXP-complete even with a known model \cite{bernstein2002complexity}), we first introduce an approach for stable single-task MARL. This enables agents to learn coordination, while also learning Q-values needed for computation of a unified MT-MARL policy.

%

\subsubsection{Decentralized Hysteretic Deep Recurrent Q-Networks (Dec-HDRQNs)}
Due to partial observability and local non-stationarity, model-based Dec-POMDP MARL is extremely challenging \cite{banerjee2012sample}. Our approach is model-free and decentralized, learning Q-values for each agent. In contrast to policy tables or FSCs, Q-values are amenable to the multi-task distillation process as they inherently measure quality of all actions, rather than just the optimal action.

Overly-optimistic MARL approaches (e.g., Distributed Q-learning \cite{lauer2000algorithm}) completely ignore low returns, which are assumed to be caused by teammates' exploratory actions. This causes severe overestimation of Q-values in stochastic domains. Hysteretic Q-learning \cite{matignon2007hysteretic}, instead, uses the insight that low returns may also be caused by domain stochasticity, which should not be ignored. This approach uses two learning rates: nominal learning rate, $\alpha$, is used when the TD-error is non-negative; a smaller learning rate, $\beta$, is used otherwise (where $0 < \beta < \alpha < 1$). The result is hysteresis (lag) of Q-value degradation for actions associated with positive past experiences that occurred due to successful cooperation. Agents are, therefore, robust against negative learning due to teammate exploration and concurrent actions. Notably, unlike Distributed Q-learning, Hysteretic Q-learning permits eventual degradation of Q-values that were overestimated due to outcomes unrelated to their associated action.

Hysteretic Q-learning has enjoyed a strong empirical track record in fully-observable MARL \cite{xu2012multiagent,matignon2012independent,barbalios2014robust}, exhibiting similar performance as more complex approaches. Encouraged by these results, we introduce Decentralized Hysteretic Deep Recurrent Q-Networks (Dec-HDRQNs) for partially-observable domains. This approach exploits the robustness of hysteresis to non-stationarity and alter-exploration, in addition to the representational power and memory-based decision making of DRQNs. As later demonstrated, Dec-HDRQN is well-suited to Dec-POMDP MARL, as opposed to non-hysteretic Dec-DRQN. 






\subsubsection{Concurrent Experience Replay Trajectories (CERTs)}
Experience replay (sampling a memory bank of experience tuples $\langle s,a,r,s' \rangle$ for TD learning) was first introduced by \citet{lin1992self} and recently shown to be crucial for stable deep Q-learning \cite{mnih2015human}. 
With experience replay, sampling cost is reduced as multiple TD updates can be conducted using each sample, enabling rapid Q-value propagation to preceding states \emph{without} additional environmental interactions. Experience replay also breaks temporal correlations of samples used for Q-value updates---crucial for reducing generalization error, as the stochastic optimization algorithms used for training DQNs typically assume i.i.d. data \cite{bengio2012practical}. 

\begin{figure}[t]
	\centering
	\begin{subfigure}[t]{0.235\textwidth}
		\centering
		\includegraphics[width=1\linewidth,page=1]{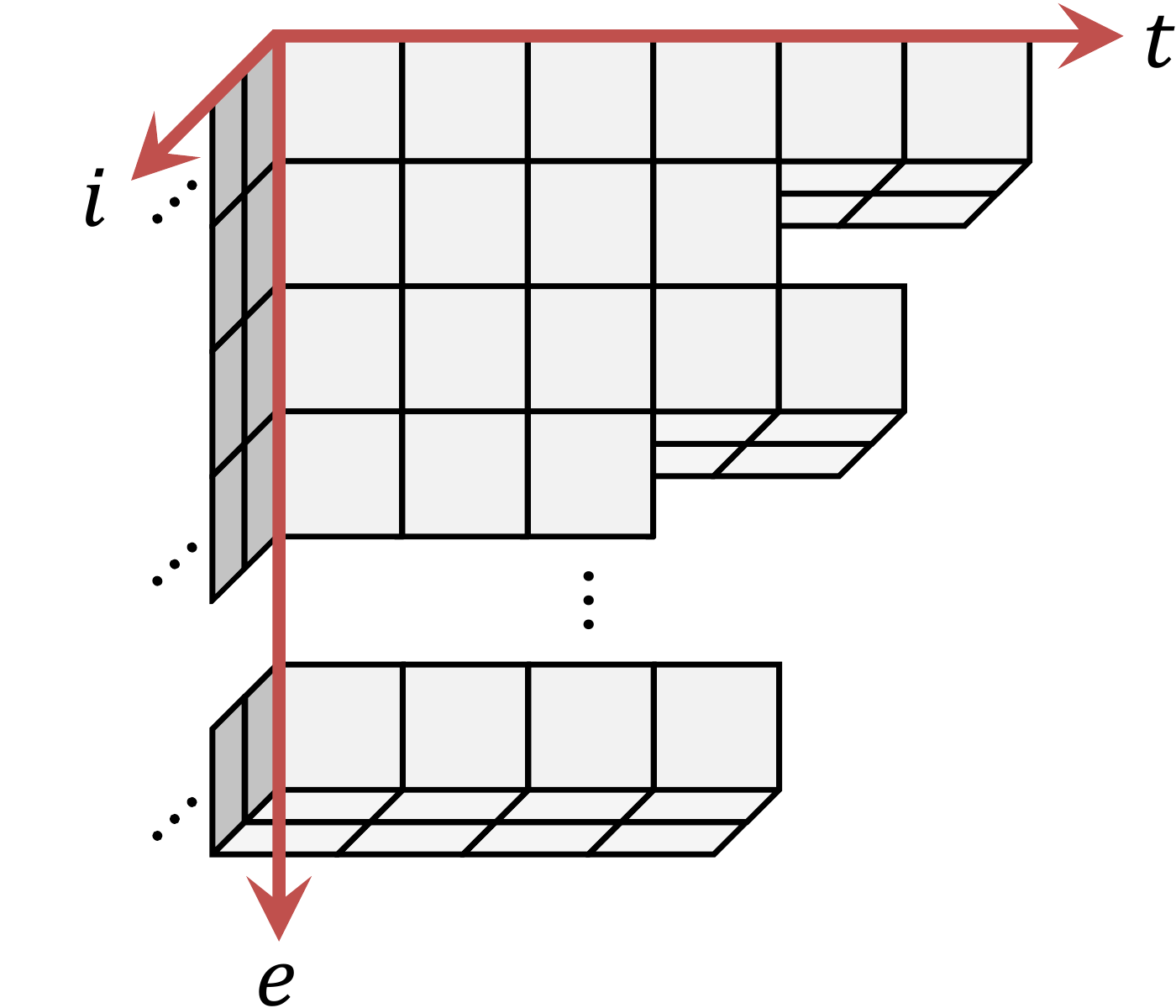}
		\caption{CERT structure.}
		\label{fig:CERT_structure}
	\end{subfigure}
	\hfill
	\begin{subfigure}[t]{0.235\textwidth}
		\centering
		\includegraphics[width=1\linewidth,page=2]{./figs/CERT.pdf}
		\caption{CERT minibatches.}
		\label{fig:CERT_minibatches}
	\end{subfigure}
	\caption{Concurrent training samples for MARL. Each cube signifies an experience tuple $\langle o_t^{(i)}, a_t^{(i)}, r_t, o_{t+1}^{(i)} \rangle$. Axes $e$, $t$, $i$ correspond to episode, timestep, and agent indices, respectively. }
	\label{fig:certs}
\end{figure}

Despite the benefits in single-agent settings, existing MARL approaches have found it necessary to disable experience replay \cite{foerster2016learning}. This is due to the non-concurrent (and non-stationary) nature of local experiences when sampled independently for each agent, despite the agents learning concurrently. A contributing factor is that inter-agent desynchronization of experiences compounds the prevalence of earlier-mentioned shadowed equilibria challenges, destabilizing coordination. As a motivating example, consider a 2 agent game where $\mathcal{A}^{(1)} = \mathcal{A}^{(2)} =\{a_1,a_2\}$. Let there be two optimal joint actions: $\langle a_1, a_1 \rangle$ and $\langle a_2, a_2 \rangle$ (e.g., only these joint actions have positive, equal reward). Given \emph{independent} experience samples for each agent, the first agent may learn action $a_1$ as optimal, whereas the second agent learns $a_2$, resulting in arbitrarily poor joint action $\langle a_1, a_2\rangle$. This motivates a need for \emph{concurrent} (synchronized) sampling of experiences across the team in MARL settings.  Concurrent experiences induce correlations in local policy updates, so that given existence of multiple equilibria, agents tend to converge to the same one. Thus, we introduce Concurrent Experience Replay Trajectories (CERTs), visualized in \cref{fig:CERT_structure}. During execution of each learning episode $e \in \mathbb{N}^{+}$, each agent $i$ collects experience tuple $\langle o_t^{(i)}, a_t^{(i)}, r_t, o_{t+1}^{(i)} \rangle$ at timestep $t$, where $o_t$, $a_t$, and $r_t$ are current observation, action, and reward, and $o_{t+1}$ is the subsequent observation. \cref{fig:CERT_structure} visualizes each experience tuple as a cube. Experiences in each episode are stored in a sequence (along time axis $t$ of \cref{fig:CERT_structure}), as Dec-HDRQN assumes an underlying RNN architecture that necessitates sequential samples for each training iteration. Importantly, as all agents are aware of timestep $t$ and episode $e$, they store their experiences concurrently (along agent index axis $i$ of \cref{fig:CERT_structure}). Upon episode termination, a new sequence is initiated (a new row along episode axis $e$ of \cref{fig:CERT_structure}). No restrictions are imposed on terminal conditions (i.e., varying trajectory lengths are permitted along axis $t$ of \cref{fig:CERT_structure}). CERTs are a first-in first-out circular queue along the episode axis $e$, such that old episodes are eventually discarded. 

\subsubsection{Training Dec-HDRQNs using CERTs}
Each agent $i$ maintains DRQN $Q^{(i)}(o_t^{(i)},h_{t-1}^{(i)},a^{(i)};\theta^{(i)})$, where $o_t^{(i)}$ is the latest local observation, $h^{(i)}_{t-1}$ is the RNN hidden state, $a^{(i)}$ is the action, and $\theta^{(i)}$ are the local DRQN parameters. DRQNs are trained on experience sequences (traces) with \emph{tracelength} $\tau$. \Cref{fig:CERT_minibatches} visualizes the minibatch sampling procedure for training, with $\tau=4$. In each training iteration, agents first sample a \emph{concurrent} minibatch of episodes. All agents' sampled traces have the same starting timesteps (i.e., are coincident along agent axis $i$ in \cref{fig:CERT_minibatches}). Guaranteed concurrent sampling merely requires a one-time (offline) consensus of agents' random number generator seeds prior to initiating learning. This ensures our approach is fully decentralized and assumes no explicit communication, even during learning. \cref{fig:CERT_minibatches} shows a minibatch of 3 episodes, $e$, sampled in red. To train DRQNs, \citet{hausknecht2015deep} suggest randomly sampling a timestep within each episode, and training using $\tau$ backward steps. However, this imposes a bias where experiences in each episode's final $\tau$ timesteps are used in fewer recurrent updates. Instead, we propose that for each sampled episode $e$, agents sample a concurrent start timestep $t_0$ for the trace from interval $\{-\tau+1,\ldots, H_e\}$, where $H_e$ is the timestep of the episode's final experience. For example, the three sampled (red) traces in \cref{fig:CERT_minibatches} start at timesteps $+1$, $-1$, and $+2$, respectively. This ensures all experiences have equal probability of being used in updates, which we found especially critical for fast training on tasks with only terminal rewards.

Sampled traces sometimes contain elements outside the episode interval (indicated as $\varnothing$ in \cref{fig:CERT_minibatches}). We discard $\varnothing$ experiences and zero-pad the suffix of associated traces (to ensure all traces have equal length $\tau$, enabling seamless use of fixed-length minibatch optimizers in standard deep learning libraries). Suffix (rather than prefix) padding ensures RNN internal states of non-$\varnothing$ samples are unaffected. In training iteration $j$, agent $i$ uses the sampling procedure to collect a minibatch of traces from CERT memory $\mathcal{M}^{(i)}$,  
\begin{align}\label{eq:CERT_minibatch}
	\mathcal{B} = \{\langle &\langle o_{t_0}^{b}, a_{t_0}^{b}, r_{t_0}^{b}, o_{t_0+1}^{b} \rangle, \ldots,\\ &\langle o_{t_0+\tau-1}^{b}, a_{t_0+\tau-1}^{b}, r_{t_0+\tau-1}^{b}, o_{t_0+\tau}^{b} \rangle\rangle\}_{b=\{1,\ldots,B\}}\nonumber,
\end{align}
where $t_0$ is the start timestep for each trace, $b$ is trace index, and $B$ is number of traces (minibatch size).\footnote{For notational simplicity, agent superscripts $(i)$ are excluded from local experiences $\langle o^{(i)}, a^{(i)}, r, o'^{(i)} \rangle$ in \cref{eq:CERT_minibatch,eq:DRQN_targets,eq:DRQN_loss,eq:distillation_loss}.} Each trace $b$ is used to calculate a corresponding sequence of target values, 
\begin{equation}\label{eq:DRQN_targets}
		\{\langle \langle y_{t_0}^{b} \rangle, \ldots, \langle y_{t_0+\tau-1}^{b}\rangle\rangle\}_{b=\{1,\ldots,B\}},
\end{equation}
where $y_t^{b} = r_t^{b} + \gamma \max_{a'}Q(o_{t+1}^{b},h_{t}^{b},a';\hat{\theta}^{(i)}_{j})$. Target network parameters $\hat{\theta}^{(i)}_{j}$ are updated less frequently, for stable learning \cite{mnih2015human}. Loss over all traces is,
\begin{equation}\label{eq:DRQN_loss}
	\thinmuskip=0mu 
	L_j(\theta_{j}^{(i)}) = \mathbb{E}_{(o_{t}^{b},a_t^{b},r_t^{b},o_{t+1}^{b})\sim \mathcal{M}^{(i)}} \lbrack(\delta_t^{b})^{2}\rbrack,
\end{equation}
where $\delta_t^{b} = y_t^{b}-Q(o_t^{b},h_{t-1}^{b},a_t^{b};\theta_{j}^{(i)})$. Loss contributions of suffix $\varnothing$-padding elements are masked out. Parameters are updated via gradient descent on \cref{eq:DRQN_loss}, with the caveat of hysteretic learning rates $0 < \beta < \alpha < 1$, where learning rate $\alpha$ is used if $\delta_t^b \geq 0$, and $\beta$ is used otherwise.

\subsection{Phase II: Dec-POMDP MT-MARL}
Following task specialization, the second phase involves distillation of each agent's set of DRQNs into a unified DRQN that performs well in all tasks without explicit provision of task ID. Using DRQNs, our approach extends the single-agent, fully-observable MTRL method proposed by \citet{rusu2015policy} to Dec-POMDP MT-MARL. Specifically, once Dec-HDRQN specialization is conducted for each task, multi-task learning can be treated as a regression problem over Q-values. During multi-task learning, our approach iteratively conducts data collection and regression. 

For data collection, agents use each specialized DRQN (from Phase I) to execute actions in corresponding tasks, resulting in a set of regression CERTs $\{\mathcal{M}_R\}$ (one per task), each containing sequences of \emph{regression experiences} $\langle o_t^{(i)}, Q_t^{(i)} \rangle$, where $Q_t^{(i)} = Q_t^{(i)}(\vec{o_t}^{(i)};\theta^{(i)})$ is the specialized DRQN's Q-value vector for agent $i$ at timestep $t$.
Supervised learning of Q-values is then conducted. Each agent samples experiences from its local regression CERTs to train a \emph{single} distilled DRQN with parameters $\theta_R^{(i)}$. Given a minibatch of regression experience traces $\mathcal{B}_R = \{\langle \langle o_{t_0}^{b}, Q_{t_0}^{b} \rangle, \ldots, \langle o_{t_0+\tau-1}^{b}, Q_{t_0+\tau-1}^{b} \rangle\rangle\}_{b=\{1,\ldots,B\}}$, the following tempered Kullback-Leibler (KL) divergence loss is minimized for each agent,
\begin{align}\label{eq:distillation_loss}
	&L_{KL}(\mathcal{B}_R,\theta_R^{(i)}; T)\\
	 &\!\!= \mathbb{E}_{(o_{t}^{b},Q_{t}^{b})\sim \{\mathcal{M_R}^{(i)}\}}\!\!\! \sum_{a=1}^{|\mathcal{A}^{(i)}|}\!\! \text{softmax}_a(\frac{Q_{t}^{b}}{T})\ln\frac{\text{softmax}_a(\frac{Q_{t}^{b}}{T})}{\text{softmax}_a(Q_{t,R}^{b})}\nonumber,
\end{align}
where $Q_{t,R}^{b} = Q_{t,R}^{b}(\vec{o_t}^{b};\theta_R^{(i)})$ is the vector of action-values predicted by distilled DRQN given the same input as the specialized DRQN, $T$ is the softmax temperature, and $\text{softmax}_a$ refers to the $a$-th element of the softmax output. The motivation behind loss function \eqref{eq:distillation_loss} is that low temperatures ($0<T<1$) lead to sharpening of specialized DRQN action-values, $Q_{t}^{b}$, ensuring that the distilled DRQN ultimately chooses similar actions as the specialized policy it was trained on. We refer readers to \citet{rusu2015policy} for additional analysis of the distillation loss. Note that concurrent sampling is not necessary during the distillation phase, as it is entirely supervised; CERTs are merely used for storage of the regression experiences.

\section{Evaluation}

\begin{figure*}[t]
	\centering
	\begin{subfigure}[t]{0.45\linewidth}
		\centering
		\includegraphics[width=1\linewidth]{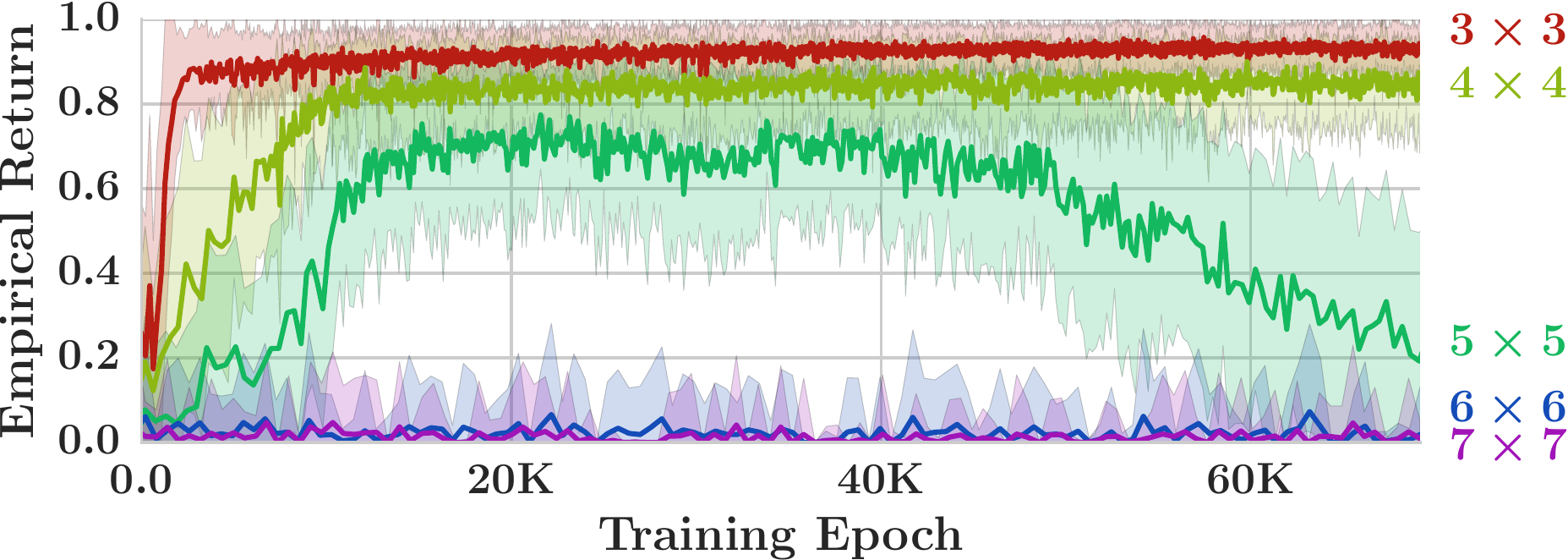}
		\caption{Learning via Dec-DRQN.}
		\label{fig:2agt_mamt_drqn_v}
	\end{subfigure}
	\hfill
	\begin{subfigure}[t]{0.45\linewidth}
		\centering
		\includegraphics[width=1\linewidth]{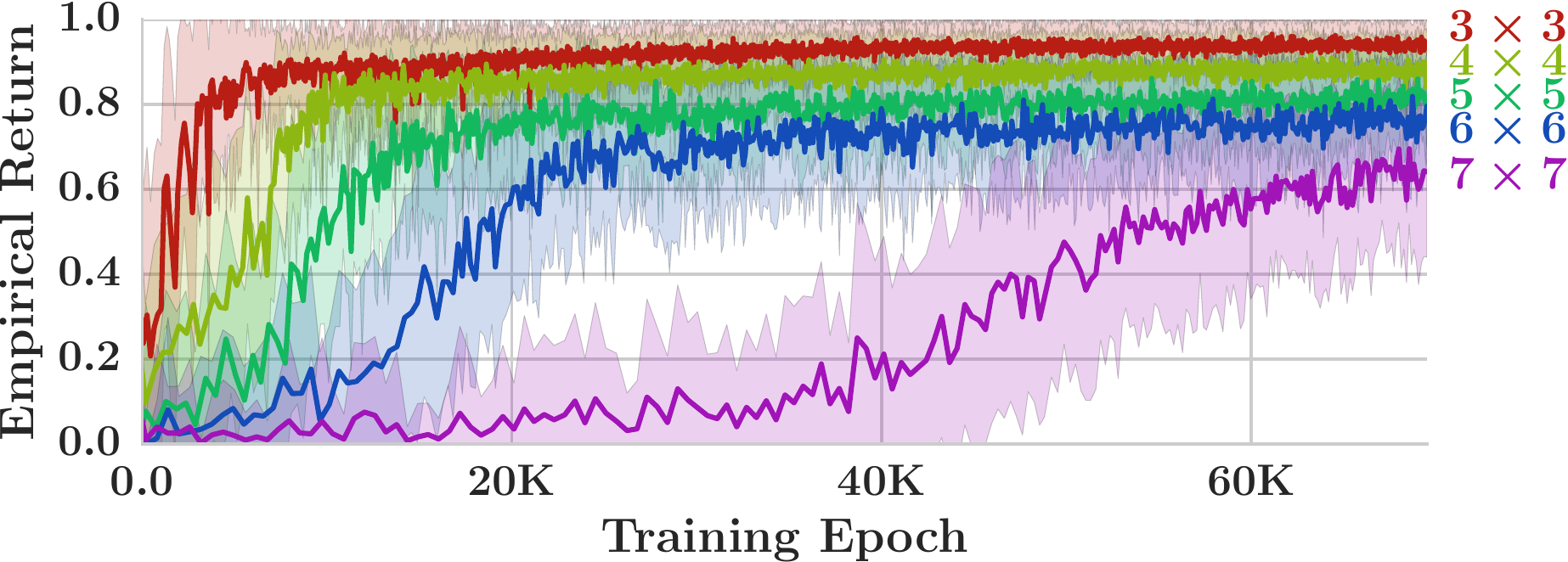}
		\caption{Learning via Dec-HDRQN (our approach).}
		\label{fig:2agt_mamt_hdrqn_v}
	\end{subfigure}
	\caption{Task specialization for MAMT domain with $n=2$ agents, $P_f = 0.3$. (a) Without hysteresis Dec-DRQN policies destabilize in the $5 \times 5$ task and fails to learn in the $6 \times 6$ and $7 \times 7$ tasks. (b) Dec-HDRQN (our approach) performs well in all tasks.}
	\label{fig:2agt_mamt_v_comparisons}
\end{figure*}

\begin{figure}[t]
	\centering
	\includegraphics[width=1\linewidth]{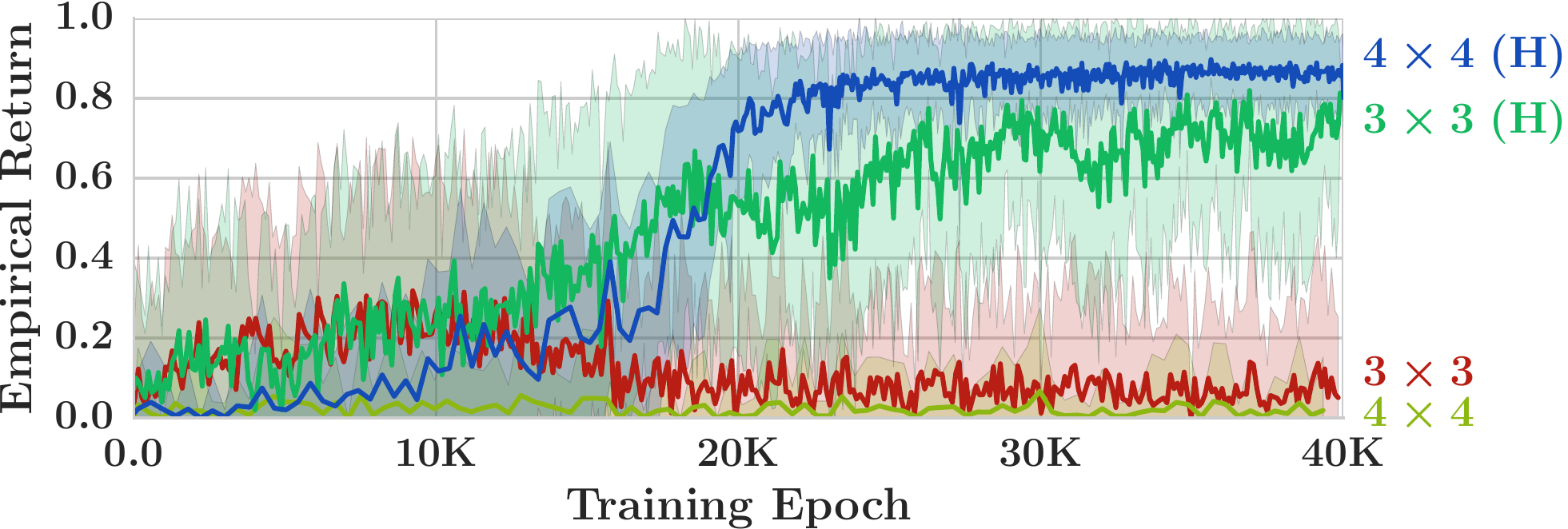}
	\caption{The advantage of hysteresis is even more pronounced for MAMT with $n=3$ agents. $P_f = 0.6$ for $3\times3$ task, and $P_f = 0.1$ for $4\times 4$ task. Dec-HDRQN indicated by (H).}
	\label{fig:3agt_v}
\end{figure}

\begin{figure}[t]
	\centering
	\includegraphics[width=1\linewidth]{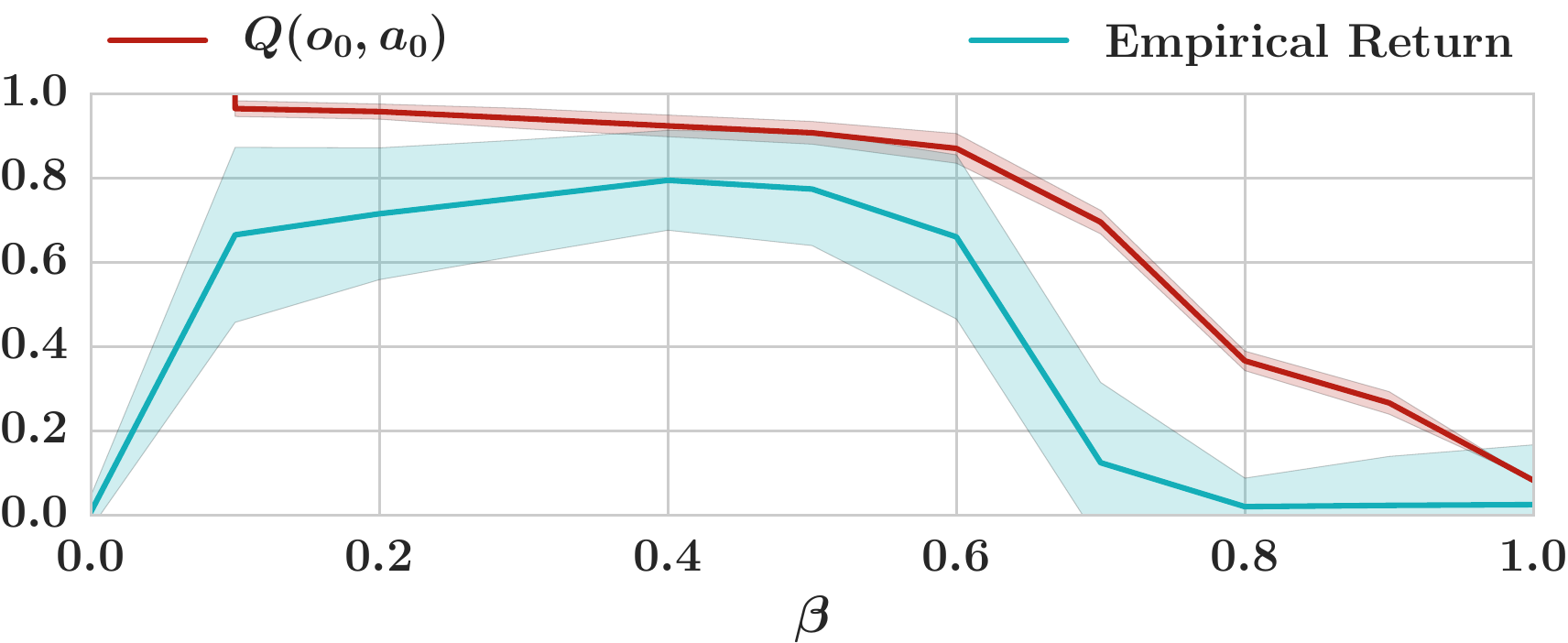}
	\caption{Dec-HDRQN sensitivity to learning rate $\beta$ ($6 \times 6$ MAMT domain, $n=2$ agents, $P_f=0.25$). Anticipated return $Q(o_0,a_0)$ upper bounds actual return due to hysteretic optimism. }
	\label{fig:beta_sensitivity}
\end{figure}

\begin{figure*}[t!]
	\centering
	\includegraphics[width=1\linewidth]{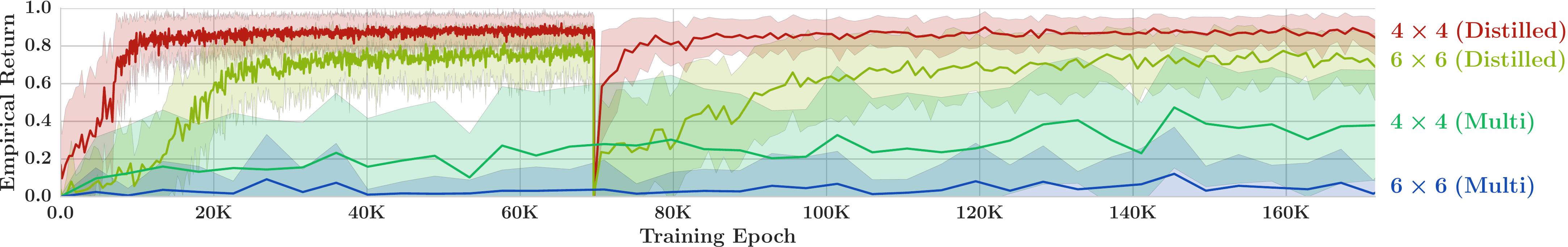}
	\caption{MT-MARL performance of the proposed Dec-HDRQN specialization/distillation approach (labeled as \emph{Distilled}) and simultaneous learning approach (labeled as \emph{Multi}). Multi-task policies for both approaches were trained on all MAMT tasks from $3 \times 3$ through $6 \times 6$. Performance shown only for $4 \times 4$ and $6 \times 6$ domains for clarity. Distilled approach shows specialization training (Phase I of approach) until 70K epochs, after which distillation is conducted (Phase II of approach). Letting the simultaneous learning approach run for up to 500K episodes did not lead to significant performance improvement. By contrast, the performance of our approach during the distillation phase (which includes task identification) is almost as good as its performance during the specialization phase.}
	\label{fig:distillation_all_combined}
\end{figure*}

\subsection{Task Specialization using Dec-HDRQN}\label{sec:expt_specialize}

We first evaluate single-task performance of the introduced Dec-HDRQN approach on a series of increasingly challenging domains. Domains are designed to support a large number of task variations, serving as a useful MT-MARL benchmarking tool. All experiments use DRQNs with 2 multi-layer perceptron (MLP) layers, an LSTM layer \cite{hochreiter1997long} with 64 memory cells, and another 2 MLP layers. MLPs have 32 hidden units each and rectified linear unit nonlinearities are used throughout, with the exception of the final (linear) layer. Experiments use $\gamma = 0.95$ and Adam optimizer \mbox{\cite{kingma2014adam}} with base learning rate 0.001. Dec-HDRQNs use hysteretic learning rate $\beta = 0.2$ to $0.4$. 
All results are reported for batches of 50 randomly-initialized episodes.

Performance is evaluated on both multi-agent single-target (MAST) and multi-agent multi-target (MAMT) capture domains, variations of the existing meeting-in-a-grid Dec-POMDP benchmark \cite{amato2009incremental}. Agents $i\in\{1,\ldots,n\}$ in an $m\times m$ toroidal grid receive $+1$ terminal reward only when they \emph{simultaneously} capture moving targets (1 target in MAST, and $n$ targets in MAMT). Each agent always observes its own location, but only sometimes observes targets' locations. Target dynamics are unknown to agents and vary across tasks. Similar to the Pong POMDP domain of \citet{hausknecht2015deep}, our domains include observation flickering: in each timestep, observations of targets are sometimes obscured, with probability $P_f$. In MAMT, each agent is assigned a unique target to capture, yet is unaware of the assignment (which also varies across tasks). Agent/target locations are randomly initialized in each episode. Actions are `move north', `south', `east', `west', and `wait', but transitions are noisy (0.1 probability of moving to an unintended adjacent cell). 

In the MAST domain, each task is specified by a unique grid size $m$; in MAMT, each task also has a unique agent-target assignment. The challenge is that agents must learn particular roles (to ensure coordination) and also discern aliased states (to ensure quick capture of targets) using local noisy observations. Tasks end after $H$ timesteps, or upon simultaneous target capture. Cardinality of local policy space for agent $i$ at timestep $t$ is $O(|\mathcal{A}^{(i)}|^{\frac{|\Omega^{(i)}|^t-1}{|\Omega^{(i)}|-1}})$ \cite{oliehoek2008optimal}, where $|A^{(i)}| = 5$, $|\Omega^{(i)}| = m^4$ for MAST, and $|\Omega^{(i)}| = m^{2(n+1)}$ for MAMT. Across all tasks, non-zero reward signals are extremely sparse, appearing in the terminal experience tuple only if targets are \emph{simultaneously} captured. Readers are referred to the supplementary material for domain visualizations.

The failure point of Dec-DRQN is first compared to Dec-HDRQN in the MAST domain with $n=2$ and $P_f = 0$ (full observability) for increasing task size, starting from $4 \times 4$. Despite the domain simplicity, Dec-DRQN fails to match Dec-HDRQN at the $8 \times 8$ mark, receiving value 0.05$\pm$0.16 in contrast to Dec-HDRQN's 0.76$\pm$0.11 (full results reported in supplementary material). 
Experiments are then scaled up to a 2 agent, 2 target MAMT domain with $P_f=0.3$. Empirical returns throughout training are shown in \cref{fig:2agt_mamt_v_comparisons}. In the MAMT tasks, a well-coordinated policy induces agents to capture targets simultaneously (yielding joint $+1$ reward). If any agent strays from this strategy during learning (e.g., while exploring), teammates receive no reward even while executing optimal local policies, leading them to deviate from learned strategies. 
Due to lack of robustness against alter-exploration/non-stationarity, the Dec-DRQN becomes unstable in the $5 \times 5$ task, and fails to learn altogether in the $6 \times 6$ and $7 \times 7$ tasks (\cref{fig:2agt_mamt_drqn_v}). Hysteresis affords Dec-HDRQN policies the stability necessary to consistently achieve agent coordination (\cref{fig:2agt_mamt_hdrqn_v}). A \emph{centralized-learning} variation of Dec-DRQN with inter-agent parameter sharing (similar to RIAL-PS in \citet{foerster2016learning}) was also tested, but was not found to improve performance (see supplementary material). These results further validate that, despite its simplicity, hysteretic learning significantly improves the stability of MARL in cooperative settings. Experiments are also conducted for the $n=3$ MAMT domain (\cref{fig:3agt_v}). This domain poses significant challenges due to reward sparsity. Even in the $4 \times 4$ task, only 0.02\% of the joint state space has a non-zero reward signal. Dec-DRQN fails to find a coordinated joint policy, receiving near-zero return after training. Dec-HDRQN successfully coordinates the 3 agents. Note the high variance in empirical return for the $3 \times 3$ task is due to flickering probability being increased to $P_f = 0.6$.

Sensitivity of Dec-HDRQN empirical performance to hysteretic learning rate $\beta$ is shown in \cref{fig:beta_sensitivity}, where lower $\beta$ corresponds to higher optimism; $\beta=0$ causes monotonic increase of approximated Q-values during learning, whereas $\beta=1$ corresponds to Dec-DRQN. Due to the optimistic assumption, anticipated returns at the initial timestep, $Q(o_0,a_0)$, overestimate true empirical return. Despite this, $\beta \in [0.1,0.6]$ consistently enables learning of a well-coordinated policy, with $\beta \in [0.4,0.5]$ achieving best performance. Readers are referred to the supplementary material for additional sensitivity analysis of convergence trends with varying $\beta$ and CERT tracelength $\tau$.







\subsection{Multi-tasking using Distilled Dec-HDRQN}
We now evaluate distillation of specialized Dec-HDRQN policies (as learned in \cref{sec:expt_specialize}) for MT-MARL. A first approach is to forgo specialization and directly learn a Dec-HDRQN using a pool of experiences from all tasks. This approach, called Multi-DQN by \citet{rusu2015policy}, is susceptible to convergence issues even in single-agent, fully-observable settings. In \cref{fig:distillation_all_combined}, we compare these approaches (where we label ours as `Distilled', and Multi-HDRQN as `Multi'). Both approaches were trained to perform multi-tasking on 2-agent MAMT tasks ranging from $3 \times 3$ to $6 \times 6$, with $P_f = 0.3$. Our distillation approach uses no task-specific MLP layers, unlike \citet{rusu2015policy}, due to our stronger assumptions on task relatedness and lack of execution-time observability of task identity.

In \cref{fig:distillation_all_combined}, our MT-MARL approach first performs Dec-HDRQN specialization training on each task for 70K epochs, and then performs distillation for 100K epochs. A grid search was conducted for temperature hyperparameter in \cref{eq:distillation_loss} ($T=0.01$ was found suitable). Note that performance is plotted only for the $4 \times 4$ and $6 \times 6$ tasks,  simply for plot clarity (see supplementary material for MT-MARL evaluation results on all tasks). Multi-HDRQN exhibits poor performance across all tasks due to the complexity involved in concurrently learning over multiple Dec-POMDPs (with partial observability, transition noise, non-stationarity, varying domain sizes, varying target dynamics, and random initializations). We experimented with larger and smaller network sizes for Multi-HDRQN, with no major difference in performance (we also include training results for 500K Multi-HDRQN iterations in the supplementary). By contrast, our proposed MT-MARL approach achieves near-nominal execution-time performance on all tasks using a single distilled policy for each agent -- despite not explicitly being provided the task identity.

\section{Contribution}
This paper introduced the first formulation and approach for  multi-task multi-agent reinforcement learning under partial observability. 
Our approach combines hysteretic learners, DRQNs, CERTs, and distillation, demonstrably achieving multi-agent coordination using a single joint policy in a set of Dec-POMDP tasks with sparse rewards, despite not being provided task identities during execution. The parametric nature of the capture tasks used for evaluation (e.g., variations in grid size, target assignments and dynamics, sensor failure probabilities) makes them good candidates for ongoing benchmarks of multi-agent multi-task learning. Future work will investigate incorporation of skills (macro-actions) into the framework, extension to domains with heterogeneous agents, and evaluation on more complex domains with much larger numbers of tasks.



%
%

\clearpage
\section*{Acknowledgements}
The authors thank the anonymous reviewers for their insightful feedback and suggestions. This work was supported by Boeing Research \& Technology, ONR MURI Grant N000141110688 and BRC Grant N000141712072.

\bibliographystyle{icml2017}
\bibliography{./references}

\begin{thebibliography}{48}
\providecommand{\natexlab}[1]{#1}
\providecommand{\url}[1]{\texttt{#1}}
\expandafter\ifx\csname urlstyle\endcsname\relax
  \providecommand{\doi}[1]{doi: #1}\else
  \providecommand{\doi}{doi: \begingroup \urlstyle{rm}\Url}\fi

\bibitem[Amato et~al.(2009)Amato, Dibangoye, and
  Zilberstein]{amato2009incremental}
Amato, Christopher, Dibangoye, Jilles~Steeve, and Zilberstein, Shlomo.
\newblock Incremental policy generation for finite-horizon {DEC-POMDPs}.
\newblock In \emph{ICAPS}, 2009.

\bibitem[Banerjee et~al.(2012)Banerjee, Lyle, Kraemer, and
  Yellamraju]{banerjee2012sample}
Banerjee, Bikramjit, Lyle, Jeremy, Kraemer, Landon, and Yellamraju, Rajesh.
\newblock Sample bounded distributed reinforcement learning for decentralized
  {POMDPs}.
\newblock In \emph{AAAI}, 2012.

\bibitem[Barbalios \& Tzionas(2014)Barbalios and Tzionas]{barbalios2014robust}
Barbalios, Nikos and Tzionas, Panagiotis.
\newblock A robust approach for multi-agent natural resource allocation based
  on stochastic optimization algorithms.
\newblock \emph{Applied Soft Computing}, 18:\penalty0 12--24, 2014.

\bibitem[Bengio(2012)]{bengio2012practical}
Bengio, Yoshua.
\newblock Practical recommendations for gradient-based training of deep
  architectures.
\newblock In \emph{Neural networks: Tricks of the trade}, pp.\  437--478.
  Springer, 2012.

\bibitem[Bernstein et~al.(2002)Bernstein, Givan, Immerman, and
  Zilberstein]{bernstein2002complexity}
Bernstein, Daniel~S, Givan, Robert, Immerman, Neil, and Zilberstein, Shlomo.
\newblock The complexity of decentralized control of markov decision processes.
\newblock \emph{Mathematics of operations research}, 27\penalty0 (4):\penalty0
  819--840, 2002.

\bibitem[Bowling \& Veloso(2002)Bowling and Veloso]{bowling2002multiagent}
Bowling, Michael and Veloso, Manuela.
\newblock Multiagent learning using a variable learning rate.
\newblock \emph{Artificial Intelligence}, 136\penalty0 (2):\penalty0 215--250,
  2002.

\bibitem[Brunskill \& Li(2013)Brunskill and Li]{brunskill2013sample}
Brunskill, Emma and Li, Lihong.
\newblock Sample complexity of multi-task reinforcement learning.
\newblock \emph{arXiv preprint arXiv:1309.6821}, 2013.

\bibitem[Bu{\c{s}}oniu et~al.(2010)Bu{\c{s}}oniu, Babu{\v{s}}ka, and
  De~Schutter]{bucsoniu2010multi}
Bu{\c{s}}oniu, Lucian, Babu{\v{s}}ka, Robert, and De~Schutter, Bart.
\newblock Multi-agent reinforcement learning: An overview.
\newblock In \emph{Innovations in multi-agent systems and applications-1}, pp.\
   183--221. Springer, 2010.

\bibitem[Caruana(1998)]{caruana1998multitask}
Caruana, Rich.
\newblock Multitask learning.
\newblock In \emph{Learning to learn}, pp.\  95--133. Springer, 1998.

\bibitem[Claus \& Boutilier(1998)Claus and Boutilier]{claus1998dynamics}
Claus, Caroline and Boutilier, Craig.
\newblock The dynamics of reinforcement learning in cooperative multiagent
  systems.
\newblock \emph{AAAI/IAAI}, 1998:\penalty0 746--752, 1998.

\bibitem[Dutech et~al.(2001)Dutech, Buffet, and Charpillet]{Dutech01}
Dutech, Alain, Buffet, Olivier, and Charpillet, Fran\c{c}ois.
\newblock Multi-agent systems by incremental gradient reinforcement learning.
\newblock In \emph{Proc. of the International Joint Conf. on Artificial
  Intelligence}, pp.\  833--838, 2001.

\bibitem[Fern{\'a}ndez \& Veloso(2006)Fern{\'a}ndez and
  Veloso]{fernandez2006probabilistic}
Fern{\'a}ndez, Fernando and Veloso, Manuela.
\newblock Probabilistic policy reuse in a reinforcement learning agent.
\newblock In \emph{Proc. of the fifth international joint conf. on Autonomous
  agents and multiagent sys.}, pp.\  720--727. ACM, 2006.

\bibitem[Foerster et~al.(2017)Foerster, Nardelli, Farquhar, Torr, Kohli,
  Whiteson, et~al.]{foerster2017stabilising}
Foerster, Jakob, Nardelli, Nantas, Farquhar, Gregory, Torr, Philip, Kohli,
  Pushmeet, Whiteson, Shimon, et~al.
\newblock Stabilising experience replay for deep multi-agent reinforcement
  learning.
\newblock \emph{arXiv preprint arXiv:1702.08887}, 2017.

\bibitem[Foerster et~al.(2016)Foerster, Assael, de~Freitas, and
  Whiteson]{foerster2016learning}
Foerster, Jakob~N, Assael, Yannis~M, de~Freitas, Nando, and Whiteson, Shimon.
\newblock Learning to communicate to solve riddles with deep distributed
  recurrent {Q}-networks.
\newblock \emph{arXiv preprint arXiv:1602.02672}, 2016.

\bibitem[Fulda \& Ventura(2007)Fulda and Ventura]{fulda2007predicting}
Fulda, Nancy and Ventura, Dan.
\newblock Predicting and preventing coordination problems in cooperative
  {Q}-learning systems.
\newblock In \emph{IJCAI}, volume 2007, pp.\  780--785, 2007.

\bibitem[Gordon(1995)]{gordon1995stable}
Gordon, Geoffrey~J.
\newblock Stable function approximation in dynamic programming.
\newblock In \emph{Proc. of the twelfth international conf. on machine
  learning}, pp.\  261--268, 1995.

\bibitem[Hausknecht \& Stone(2015)Hausknecht and Stone]{hausknecht2015deep}
Hausknecht, Matthew and Stone, Peter.
\newblock Deep recurrent {Q}-learning for partially observable {MDPs}.
\newblock \emph{arXiv preprint arXiv:1507.06527}, 2015.

\bibitem[Hinton et~al.(2015)Hinton, Vinyals, and Dean]{hinton2015distilling}
Hinton, Geoffrey, Vinyals, Oriol, and Dean, Jeff.
\newblock Distilling the knowledge in a neural network.
\newblock \emph{arXiv preprint arXiv:1503.02531}, 2015.

\bibitem[Hochreiter \& Schmidhuber(1997)Hochreiter and
  Schmidhuber]{hochreiter1997long}
Hochreiter, Sepp and Schmidhuber, J{\"u}rgen.
\newblock Long short-term memory.
\newblock \emph{Neural comp.}, 9\penalty0 (8):\penalty0 1735--1780, 1997.

\bibitem[Kaelbling et~al.(1998)Kaelbling, Littman, and
  Cassandra]{kaelbling1998planning}
Kaelbling, Leslie~Pack, Littman, Michael~L, and Cassandra, Anthony~R.
\newblock Planning and acting in partially observable stochastic domains.
\newblock \emph{Artificial intelligence}, 101\penalty0 (1):\penalty0 99--134,
  1998.

\bibitem[Kapetanakis \& Kudenko(2002)Kapetanakis and
  Kudenko]{kapetanakis2002reinforcement}
Kapetanakis, Spiros and Kudenko, Daniel.
\newblock Reinforcement learning of coordination in cooperative multi-agent
  systems.
\newblock \emph{AAAI/IAAI}, 2002:\penalty0 326--331, 2002.

\bibitem[Kingma \& Ba(2014)Kingma and Ba]{kingma2014adam}
Kingma, Diederik and Ba, Jimmy.
\newblock Adam: A method for stochastic optimization.
\newblock \emph{arXiv preprint arXiv:1412.6980}, 2014.

\bibitem[Lauer \& Riedmiller(2000)Lauer and Riedmiller]{lauer2000algorithm}
Lauer, Martin and Riedmiller, Martin.
\newblock An algorithm for distributed reinforcement learning in cooperative
  multi-agent systems.
\newblock In \emph{Proc. of the Seventeenth International Conf. on Machine
  Learning}. Citeseer, 2000.

\bibitem[Laurent et~al.(2011)Laurent, Matignon, Fort-Piat,
  et~al.]{laurent2011world}
Laurent, Guillaume~J, Matignon, La{\"e}titia, Fort-Piat, Le, et~al.
\newblock The world of independent learners is not markovian.
\newblock \emph{International Journal of Knowledge-based and Intelligent
  Engineering Systems}, 15\penalty0 (1):\penalty0 55--64, 2011.

\bibitem[Lin(1992)]{lin1992self}
Lin, Long-Ji.
\newblock Self-improving reactive agents based on reinforcement learning,
  planning and teaching.
\newblock \emph{Machine learning}, 8\penalty0 (3-4):\penalty0 293--321, 1992.

\bibitem[Liu et~al.(2015)Liu, Amato, Liao, Carin, and How]{LiuIJCAI15}
Liu, Miao, Amato, Christopher, Liao, Xuejun, Carin, Lawrence, and How,
  Jonathan~P.
\newblock Stick-breaking policy learning in {Dec-POMDPs}.
\newblock In \emph{Proc. of the International Joint Conf. on Artificial
  Intelligence}, 2015.

\bibitem[Liu et~al.(2016)Liu, Amato, Anesta, Griffith, and How]{LiuAAAI16}
Liu, Miao, Amato, Christopher, Anesta, Emily, Griffith, J.~Daniel, and How,
  Jonathan~P.
\newblock Learning for decentralized control of multiagent systems in large
  partially observable stochastic environments.
\newblock In \emph{AAAI}, 2016.

\bibitem[Matignon et~al.(2007)Matignon, Laurent, and
  Le~Fort-Piat]{matignon2007hysteretic}
Matignon, La{\"e}titia, Laurent, Guillaume~J, and Le~Fort-Piat, Nadine.
\newblock Hysteretic {Q}-learning: an algorithm for decentralized reinforcement
  learning in cooperative multi-agent teams.
\newblock In \emph{IROS}, 2007.

\bibitem[Matignon et~al.(2012)Matignon, Laurent, and
  Le~Fort-Piat]{matignon2012independent}
Matignon, Laetitia, Laurent, Guillaume~J, and Le~Fort-Piat, Nadine.
\newblock Independent reinforcement learners in cooperative markov games: a
  survey regarding coordination problems.
\newblock \emph{The Knowledge Engineering Review}, 27\penalty0 (01):\penalty0
  1--31, 2012.

\bibitem[Mnih et~al.(2015)Mnih, Kavukcuoglu, Silver, Rusu, Veness, Bellemare,
  Graves, Riedmiller, Fidjeland, Ostrovski, et~al.]{mnih2015human}
Mnih, Volodymyr, Kavukcuoglu, Koray, Silver, David, Rusu, Andrei~A, Veness,
  Joel, Bellemare, Marc~G, Graves, Alex, Riedmiller, Martin, Fidjeland,
  Andreas~K, Ostrovski, Georg, et~al.
\newblock Human-level control through deep reinforcement learning.
\newblock \emph{Nature}, 518\penalty0 (7540):\penalty0 529--533, 2015.

\bibitem[Oliehoek \& Amato(2016)Oliehoek and Amato]{DecPOMDPBook16}
Oliehoek, Frans~A. and Amato, Christopher.
\newblock \emph{A Concise Introduction to Decentralized {POMDPs}}.
\newblock Springer, 2016.

\bibitem[Oliehoek et~al.(2008)Oliehoek, Spaan, Vlassis,
  et~al.]{oliehoek2008optimal}
Oliehoek, Frans~A, Spaan, Matthijs~TJ, Vlassis, Nikos~A, et~al.
\newblock Optimal and approximate q-value functions for decentralized {POMDPs}.
\newblock \emph{Journal of Artificial Intelligence Research (JAIR)},
  32:\penalty0 289--353, 2008.

\bibitem[Pan \& Yang(2010)Pan and Yang]{pan2010survey}
Pan, Sinno~Jialin and Yang, Qiang.
\newblock A survey on transfer learning.
\newblock \emph{IEEE Transactions on knowledge and data engineering},
  22\penalty0 (10):\penalty0 1345--1359, 2010.

\bibitem[Peshkin et~al.(2000)Peshkin, Kim, Meuleau, and
  Kaelbling]{peshkin2000learning}
Peshkin, Leonid, Kim, Kee-Eung, Meuleau, Nicolas, and Kaelbling, Leslie~Pack.
\newblock Learning to cooperate via policy search.
\newblock In \emph{Proc. of the Sixteenth conf. on Uncertainty in artificial
  intelligence}, pp.\  489--496. Morgan Kaufmann Publishers Inc., 2000.

\bibitem[Rusu et~al.(2015)Rusu, Colmenarejo, Gulcehre, Desjardins, Kirkpatrick,
  Pascanu, Mnih, Kavukcuoglu, and Hadsell]{rusu2015policy}
Rusu, Andrei~A, Colmenarejo, Sergio~Gomez, Gulcehre, Caglar, Desjardins,
  Guillaume, Kirkpatrick, James, Pascanu, Razvan, Mnih, Volodymyr, Kavukcuoglu,
  Koray, and Hadsell, Raia.
\newblock Policy distillation.
\newblock \emph{arXiv preprint arXiv:1511.06295}, 2015.

\bibitem[Sutton \& Barto(1998)Sutton and Barto]{sutton1998reinforcement}
Sutton, Richard~S and Barto, Andrew~G.
\newblock \emph{Reinforcement learning: An introduction}, volume~1.
\newblock MIT press Cambridge, 1998.

\bibitem[Tan(1993)]{tan1993multi}
Tan, Ming.
\newblock Multi-agent reinforcement learning: Independent vs. cooperative
  agents.
\newblock In \emph{Proc. of the tenth international conf. on machine learning},
  pp.\  330--337, 1993.

\bibitem[Tanaka \& Yamamura(2003)Tanaka and Yamamura]{tanaka2003multitask}
Tanaka, Fumihide and Yamamura, Masayuki.
\newblock Multitask reinforcement learning on the distribution of mdps.
\newblock In \emph{Computational Intelligence in Robotics and Automation, 2003.
  Proceedings. 2003 IEEE International Symposium on}, volume~3, pp.\
  1108--1113. IEEE, 2003.

\bibitem[Taylor et~al.(2013)Taylor, Dusparic, Galv{\'a}n-L{\'o}pez, Clarke, and
  Cahill]{taylor2013transfer}
Taylor, Adam, Dusparic, Ivana, Galv{\'a}n-L{\'o}pez, Edgar, Clarke,
  Siobh{\'a}n, and Cahill, Vinny.
\newblock Transfer learning in multi-agent systems through parallel transfer.
\newblock In \emph{Workshop on Theoretically Grounded Transfer Learning at the
  30th International Conf. on Machine Learning (Poster)}, volume~28, pp.\ ~28.
  Omnipress, 2013.

\bibitem[Taylor \& Stone(2009)Taylor and Stone]{taylor2009transfer}
Taylor, Matthew~E and Stone, Peter.
\newblock Transfer learning for reinforcement learning domains: A survey.
\newblock \emph{Journal of Machine Learning Research}, 10\penalty0
  (Jul):\penalty0 1633--1685, 2009.

\bibitem[Torrey \& Shavlik(2009)Torrey and Shavlik]{torrey2009transfer}
Torrey, Lisa and Shavlik, Jude.
\newblock Transfer learning.
\newblock \emph{Handbook of Research on Machine Learning Applications and
  Trends: Algs., Methods, and Techniques}, 1:\penalty0 242, 2009.

\bibitem[Watkins \& Dayan(1992)Watkins and Dayan]{watkins1992q}
Watkins, Christopher~JCH and Dayan, Peter.
\newblock Q-learning.
\newblock \emph{Machine learning}, 8\penalty0 (3-4):\penalty0 279--292, 1992.

\bibitem[Wierstra et~al.(2007)Wierstra, Foerster, Peters, and
  Schmidhuber]{wierstra2007solving}
Wierstra, Daan, Foerster, Alexander, Peters, Jan, and Schmidhuber, Juergen.
\newblock Solving deep memory {POMDPs} with recurrent policy gradients.
\newblock In \emph{International Conf. on Artificial Neural Networks}, pp.\
  697--706. Springer, 2007.

\bibitem[Wilson et~al.(2007)Wilson, Fern, Ray, and Tadepalli]{wilson2007multi}
Wilson, Aaron, Fern, Alan, Ray, Soumya, and Tadepalli, Prasad.
\newblock Multi-task reinforcement learning: a hierarchical bayesian approach.
\newblock In \emph{Proc. of the 24th international conf. on Machine learning},
  pp.\  1015--1022. ACM, 2007.

\bibitem[Wilson et~al.(2008)Wilson, Fern, Ray, and
  Tadepalli]{wilson2008learning}
Wilson, Aaron, Fern, Alan, Ray, Soumya, and Tadepalli, Prasad.
\newblock Learning and transferring roles in multi-agent reinforcement.
\newblock In \emph{Proc. AAAI-08 Workshop on Transfer Learning for Complex
  Tasks}, 2008.

\bibitem[Wu et~al.(2012)Wu, Zilberstein, and Chen]{wu2012rollout}
Wu, Feng, Zilberstein, Shlomo, and Chen, Xiaoping.
\newblock Rollout sampling policy iteration for decentralized {POMDPs}.
\newblock \emph{arXiv preprint arXiv:1203.3528}, 2012.

\bibitem[Wu et~al.(2013)Wu, Zilberstein, and Jennings]{Wu13}
Wu, Feng, Zilberstein, Shlomo, and Jennings, Nicholas~R.
\newblock Monte-carlo expectation maximization for decentralized {POMDPs}.
\newblock In \emph{Proc. of the International Joint Conf. on Artificial
  Intelligence}, pp.\  397--403, 2013.

\bibitem[Xu et~al.(2012)Xu, Zhang, Liu, and Ferrese]{xu2012multiagent}
Xu, Yinliang, Zhang, Wei, Liu, Wenxin, and Ferrese, Frank.
\newblock Multiagent-based reinforcement learning for optimal reactive power
  dispatch.
\newblock \emph{IEEE Transactions on Systems, Man, and Cybernetics, Part C
  (Applications and Reviews)}, 42\penalty0 (6):\penalty0 1742--1751, 2012.

\end{thebibliography}

\clearpage
\appendix

\onecolumn
\pagebreak

\begin{center}
	\textbf{\large Supplemental: Deep Decentralized Multi-task Multi-agent RL under Partial Observability}
\end{center}

\emph{The following provides additional empirical results. Some plots are reproduced from the main text to ease comparisons.}
\vspace{-10pt}

\section{Multi-agent Multi-target (MAMT) Domain Overview}
\begin{figure*}[h]
	\centering
	\begin{subfigure}[t]{0.3\textwidth}
		\centering
		\includegraphics[page=1,width=1\linewidth]{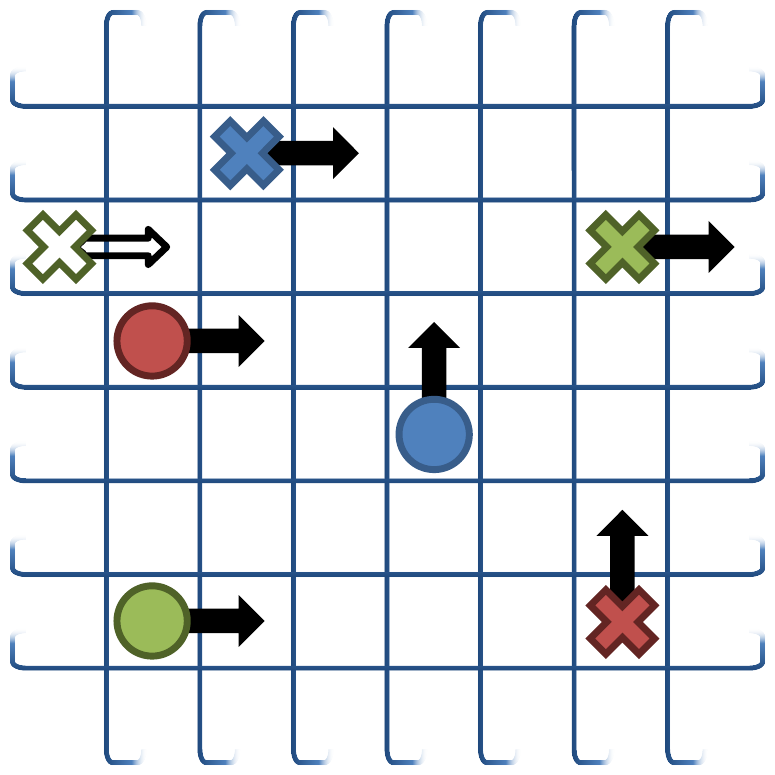}
		\caption{Agents must learn inherent toroidal transition dynamics in the domain for fast target capture (e.g., see green target).}
		\label{fig:toroidal_domain_init}
	\end{subfigure}
	\hfill
		\begin{subfigure}[t]{0.3\textwidth}
		\centering
		\includegraphics[page=2,width=1\linewidth]{./figs/toroidal_domain.pdf}
		\caption{In MAMT tasks, no reward is given to the team above, despite two agents successfully capturing their targets.}
		\label{fig:toroidal_domain_bad}
	\end{subfigure}
	\hfill
	\begin{subfigure}[t]{0.3\textwidth}
	\centering
	\includegraphics[page=3,width=1\linewidth]{./figs/toroidal_domain.pdf}
	\caption{Example successful simultaneous capture scenario, where the team is given $+1$ reward.}
	\label{fig:toroidal_domain_good}
	\end{subfigure}
\caption{Visualization of MAMT domain. Agents and targets operate on a toroidal $m \times m$ gridworld. Each agent (circle) is assigned a unique target (cross) to capture, but does not observe its assigned target ID. Targets' states are fully occluded at each timestep with probability $P_f$. Despite the simplicity of gridworld transitions, reward sparsity makes this an especially challenging task. During both learning and execution, the team receives no reward unless all targets are captured \emph{simultaneously} by their corresponding agents.}
\end{figure*}

\section{Empirical Results: Learning on Multi-agent Single-Target (MAST) Domain}
Multi-agent Single-target (MAST) domain results for Dec-DRQN and Dec-HDRQN, with 2 agents and $P_f = 0.0$ (observation flickering disabled). These results mainly illustrate that Dec-DRQN sometimes has \emph{some} empirical success in low-noise domains with small state-space. Note that in the $8 \times 8$ task, Dec-HDRQN significantly outperforms Dec-DRQN, which converges to a sub-optimal policy despite domain simplicity.

\begin{figure*}[h]
	\centering
	\begin{subfigure}[t]{0.48\textwidth}
		\centering
		\includegraphics[width=1\linewidth]{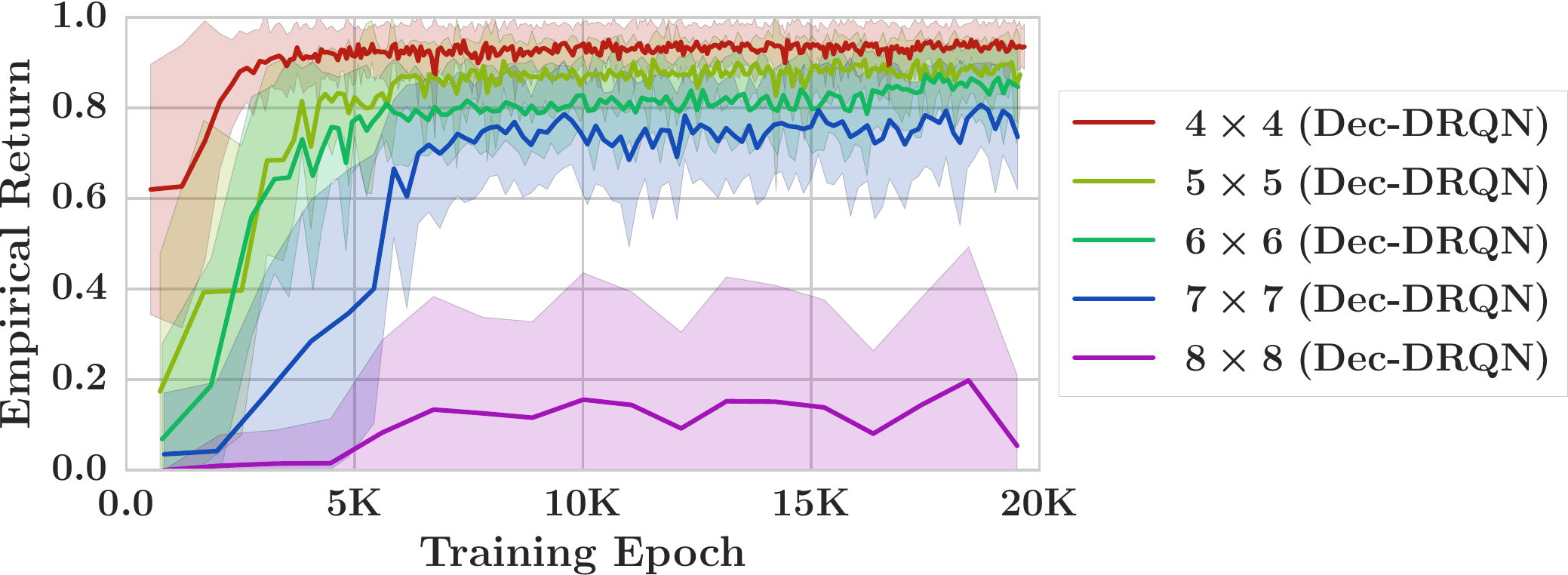}
		\caption{Dec-DRQN empirical returns during training.}
		\label{fig:2agt_mast_drqn_v}
	\end{subfigure}
	\hfill
	\begin{subfigure}[t]{0.48\textwidth}
		\centering
		\includegraphics[width=1\linewidth]{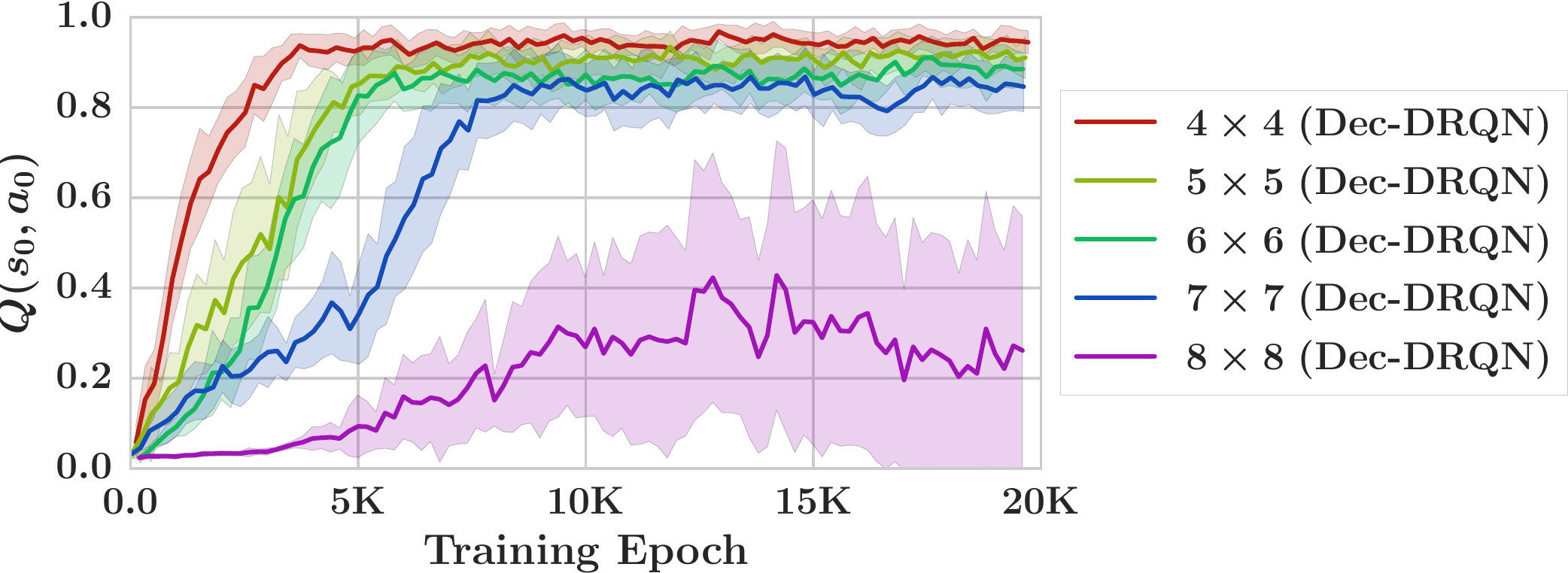}
		\caption{Dec-DRQN anticipated values during training. }
		\label{fig:2agt_mast_drqn_q}
	\end{subfigure}
	\begin{subfigure}[t]{0.48\textwidth}
		\centering
		\includegraphics[width=1\linewidth]{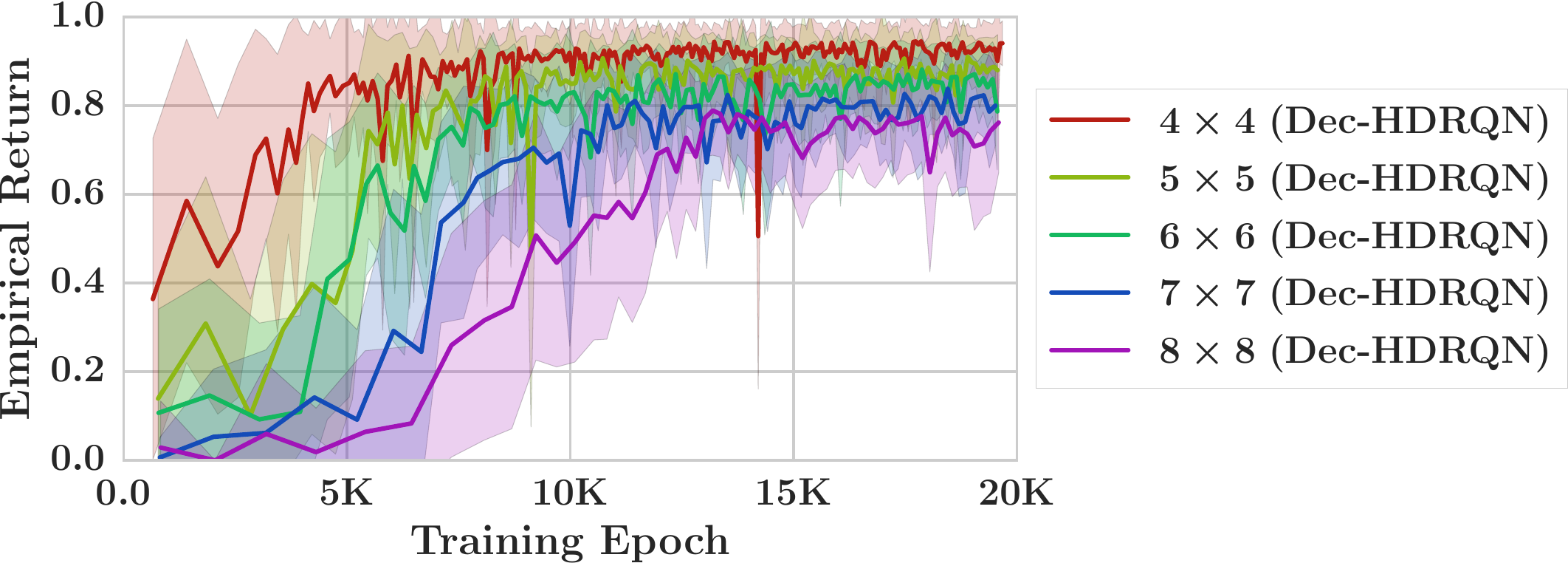}
		\caption{Dec-HDRQN empirical returns during training.}
		\label{fig:2agt_mast_hdrqn_v}
	\end{subfigure}
	\hfill
	\begin{subfigure}[t]{0.48\textwidth}
		\centering
		\includegraphics[width=1\linewidth]{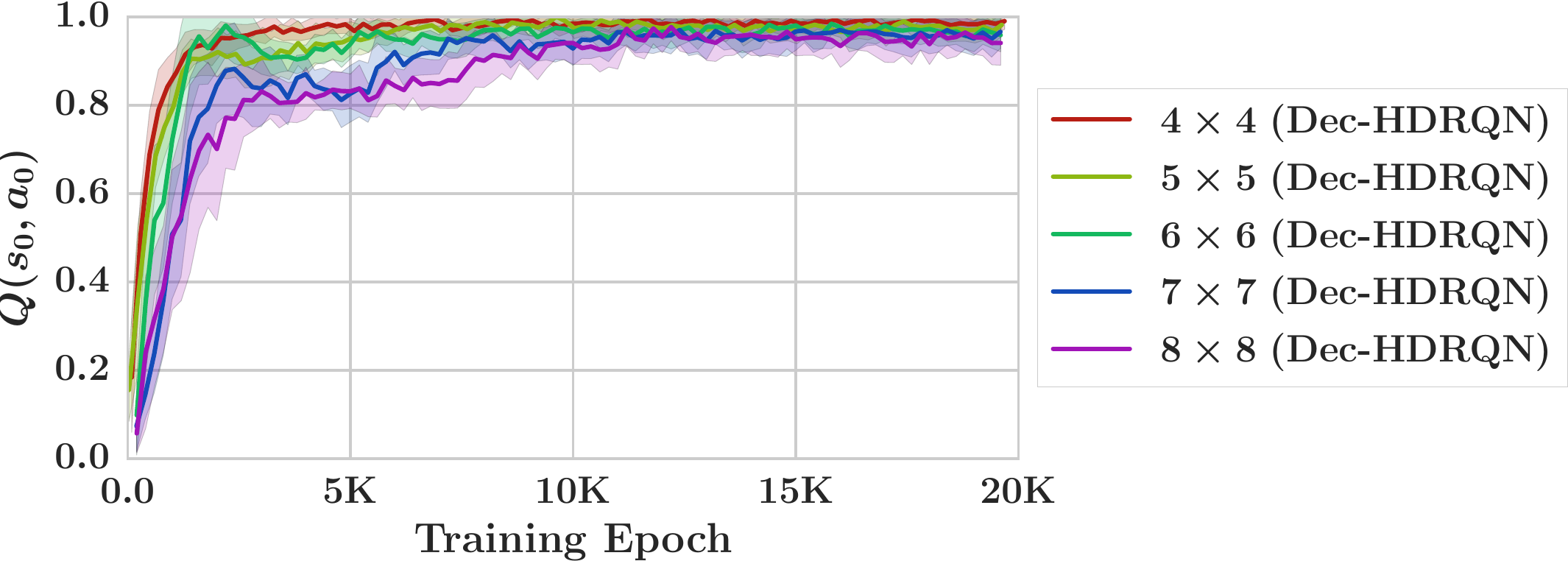}
		\caption{Dec-HDRQN anticipated values during training. }
		\label{fig:2agt_mast_hdrqn_q}
	\end{subfigure}
	\caption{Multi-agent Single-target (MAST) domain results for Dec-DRQN and Dec-HDRQN, with 2 agents and $P_f = 0.0$ (observation flickering disabled). All plots conducted (at each training epoch) for a batch of 50 randomly-initialized episodes. Anticipated value plots (on right) were plotted for the exact starting states and actions undertaken for the episodes used in the plots on the left.}
	\label{fig:mast_plots_comparisons}
\end{figure*}

\clearpage
\section{Empirical Results: Learning on MAMT Domain}

Multi-agent Single-target (MAMT) domain results, with 2 agents and $P_f = 0.3$ (observation flickering disabled). We also evaluated performance of inter-agent parameter sharing (a centralized approach) in Dec-DRQN (which we called Dec-DRQN-PS). Additionally, performance of a Double-DQN was deemed to have negligible impacts (Dec-DDRQN).


\begin{figure*}[h]
	\centering
	\begin{subfigure}[t]{0.48\textwidth}
		\centering
		\includegraphics[width=1\linewidth]{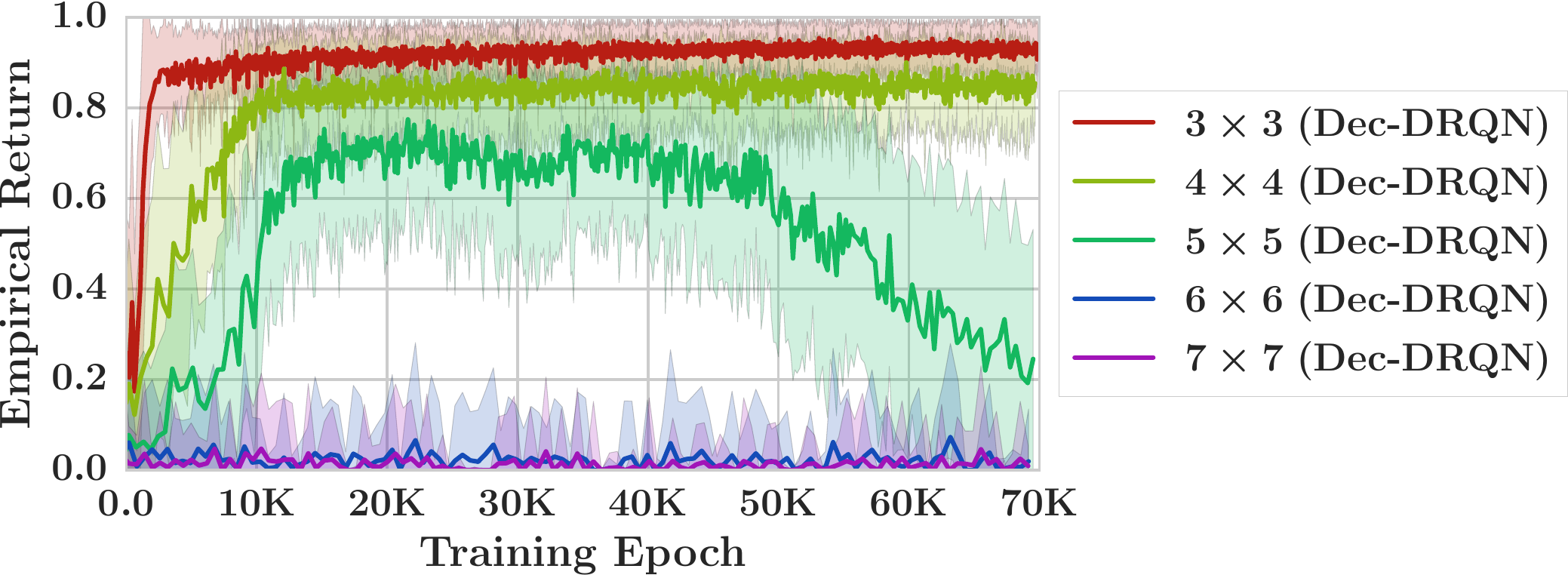}
		\caption{Dec-DRQN empirical returns during training. }
		\label{fig:2agt_mamt_drqn_v_appendix}
	\end{subfigure}
	\hfill
	\begin{subfigure}[t]{0.48\textwidth}
		\centering
		\includegraphics[width=1\linewidth]{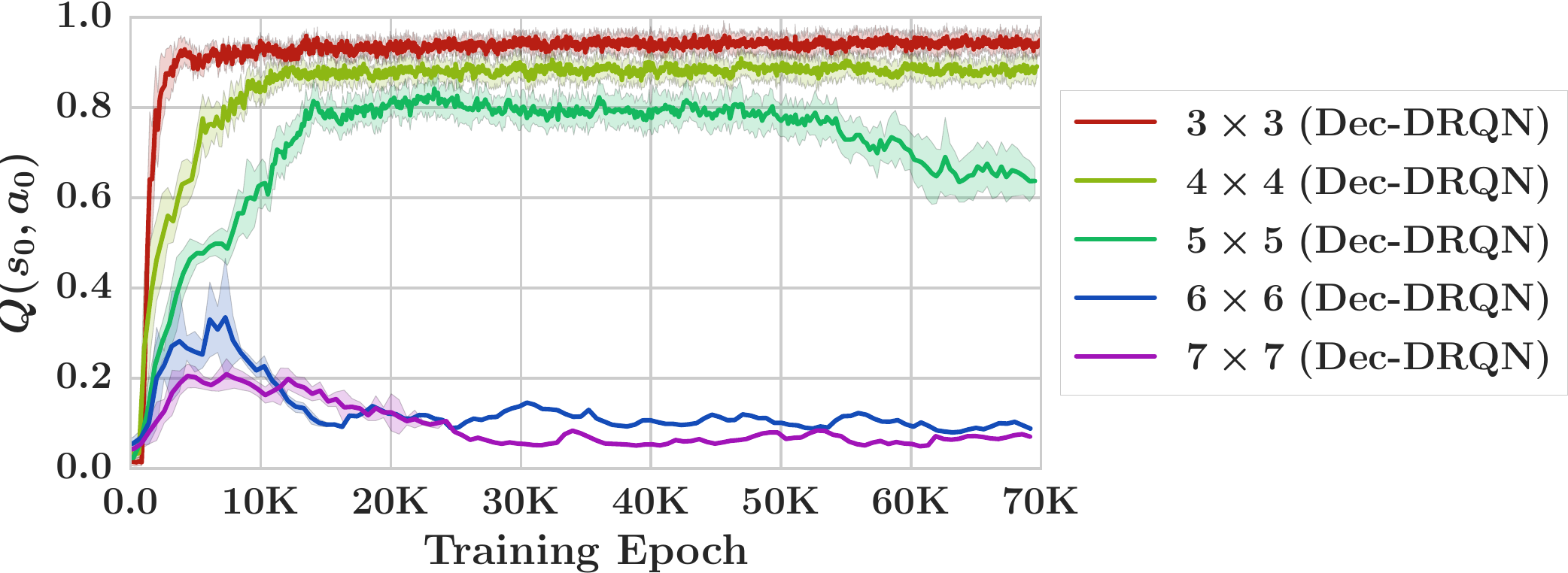}
		\caption{Dec-DRQN anticipated values during training. }
		\label{fig:2agt_mamt_drqn_q_appendix}
	\end{subfigure}
	\begin{subfigure}[t]{0.48\textwidth}
		\centering
		\includegraphics[width=1\linewidth]{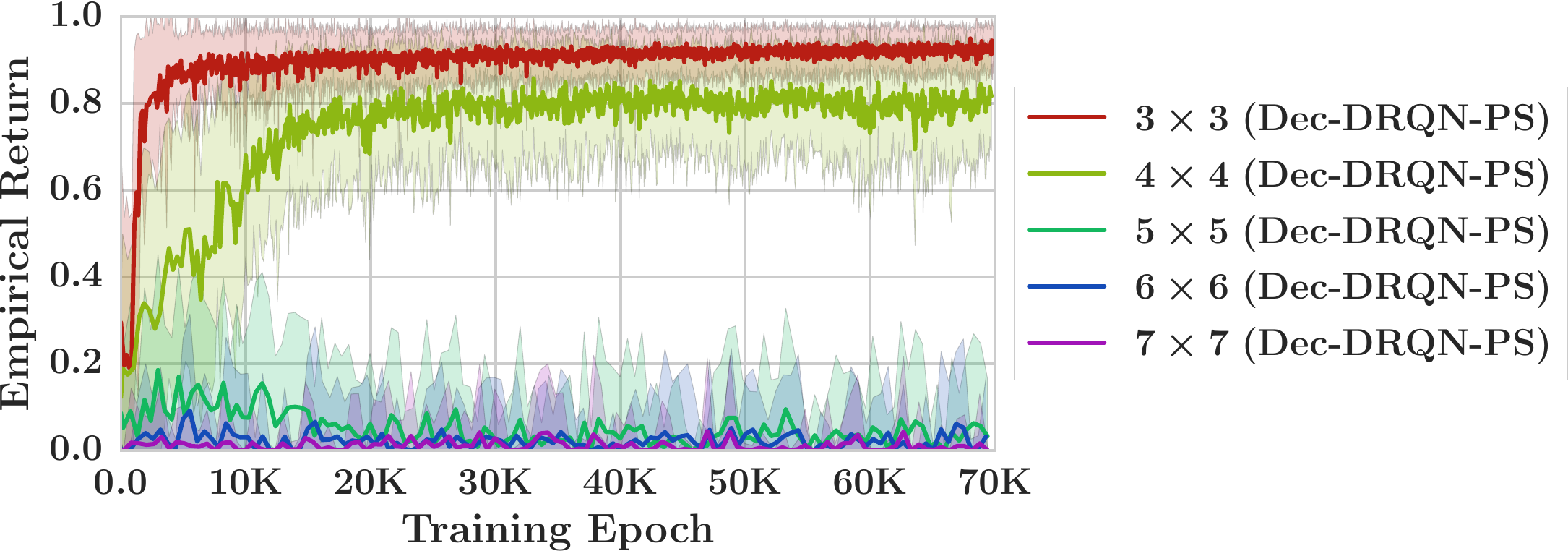}
		\caption{Dec-DRQN-PS (with parameter sharing), empirical returns during training.}
		\label{fig:2agt_mamt_drqn_ps_v_appendix}
	\end{subfigure}
	\hfill
	\begin{subfigure}[t]{0.48\textwidth}
		\centering
		\includegraphics[width=1\linewidth]{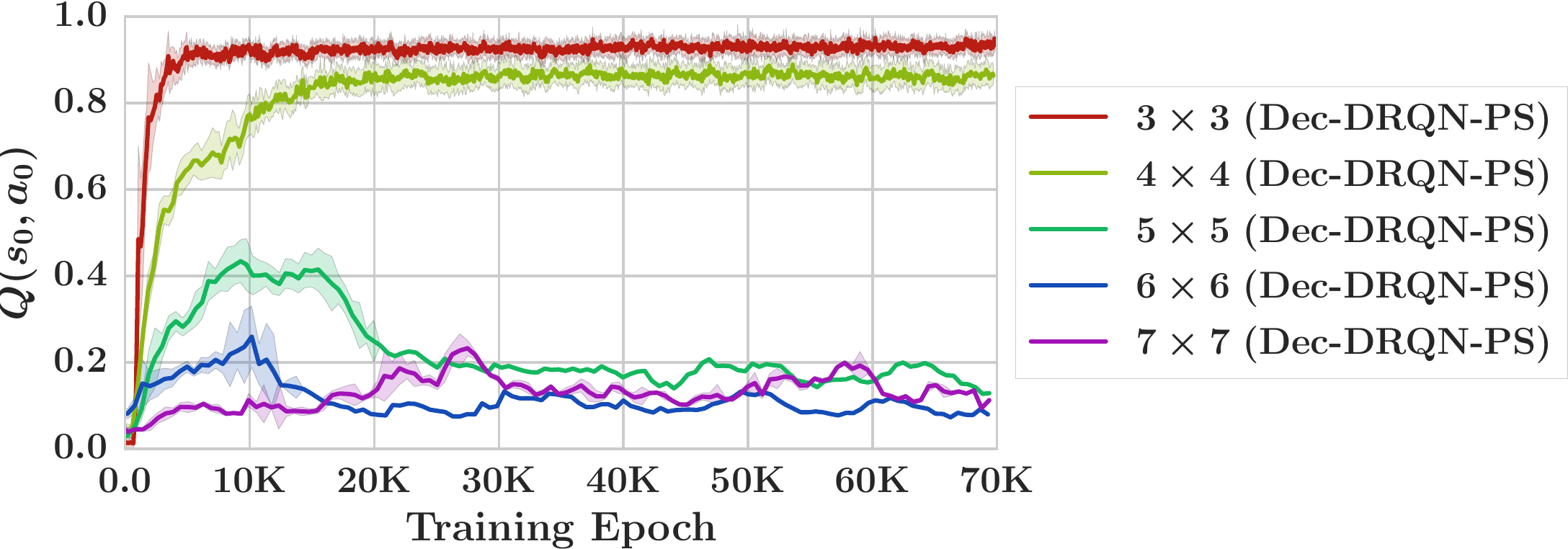}
		\caption{Dec-DRQN-PS (with parameter sharing), anticipated values during training.}
		\label{fig:2agt_mamt_drqn_ps_q_init_appendix}
	\end{subfigure}
	\begin{subfigure}[t]{0.48\textwidth}
		\centering
		\includegraphics[width=1\linewidth]{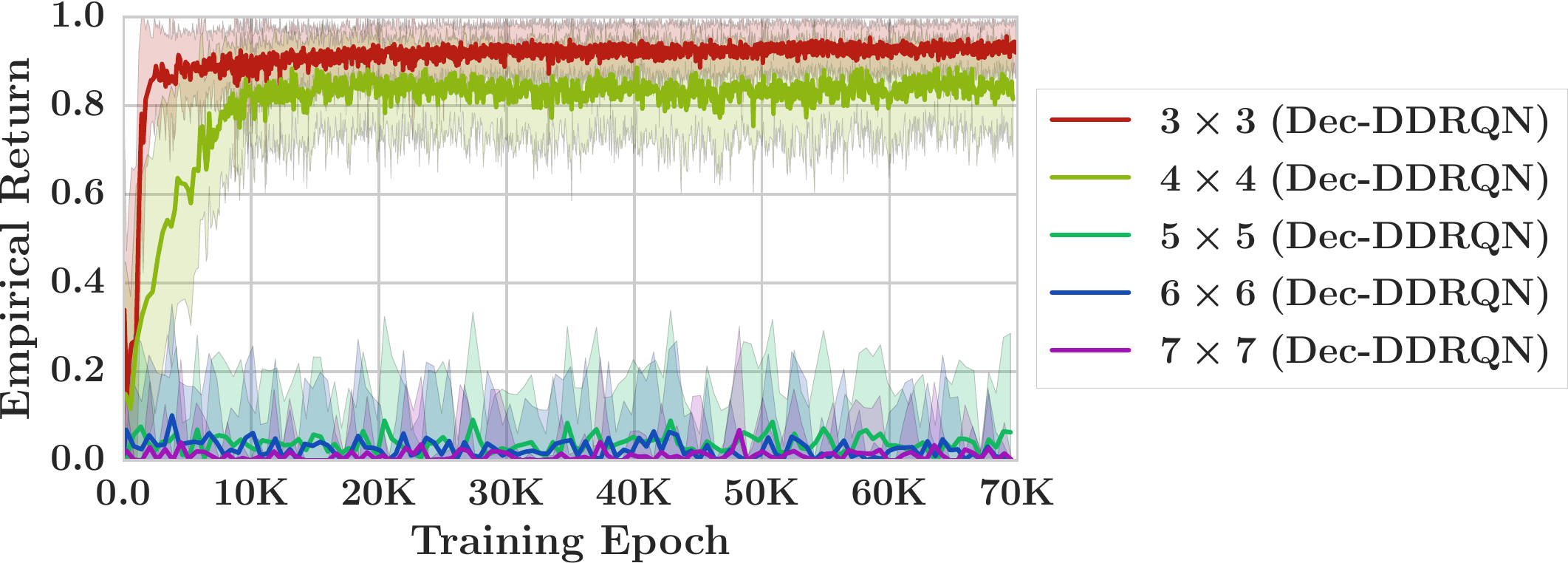}
		\caption{Dec-DDRQN (double DRQN) empirical returns during training. }
		\label{fig:2agt_mamt_double_drqn_v_appendix}
	\end{subfigure}
	\hfill
	\begin{subfigure}[t]{0.48\textwidth}
		\centering
		\includegraphics[width=1\linewidth]{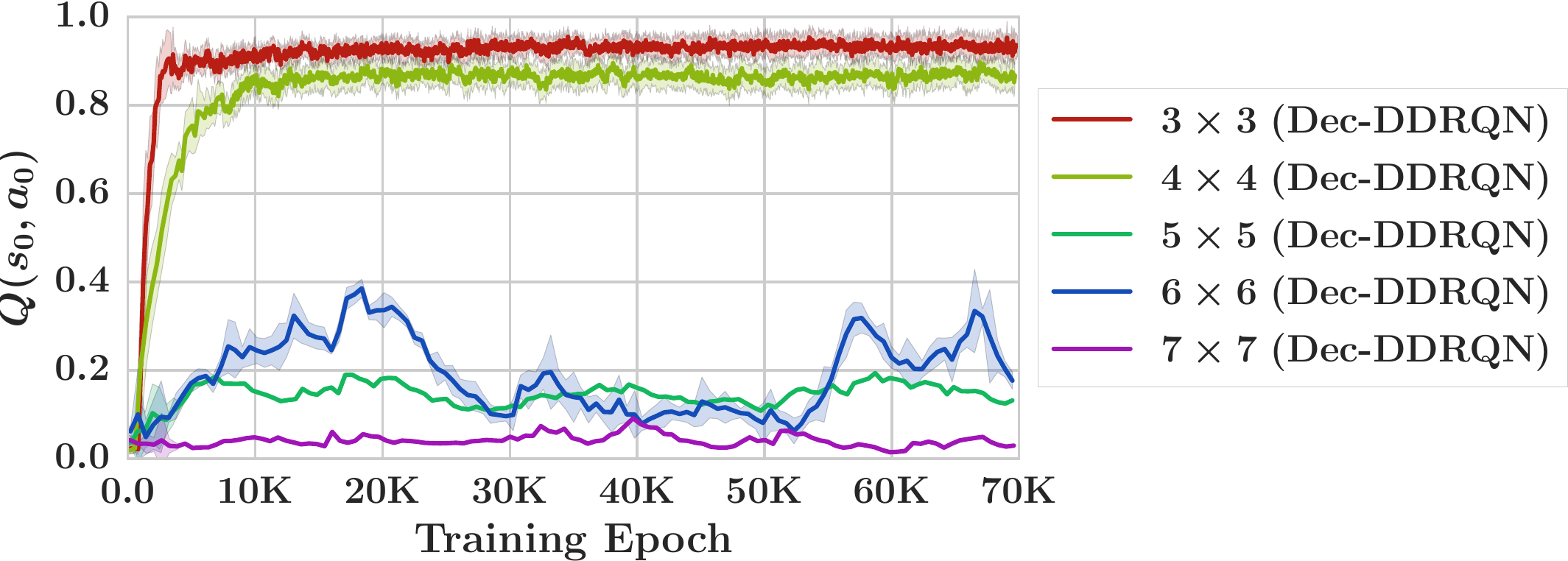}
		\caption{Dec-DDRQN (double DRQN) anticipated values during training. }
		\label{fig:2agt_mamt_double_drqn_q_init_appendix}
	\end{subfigure}
	\begin{subfigure}[t]{0.48\textwidth}
		\centering
		\includegraphics[width=1\linewidth]{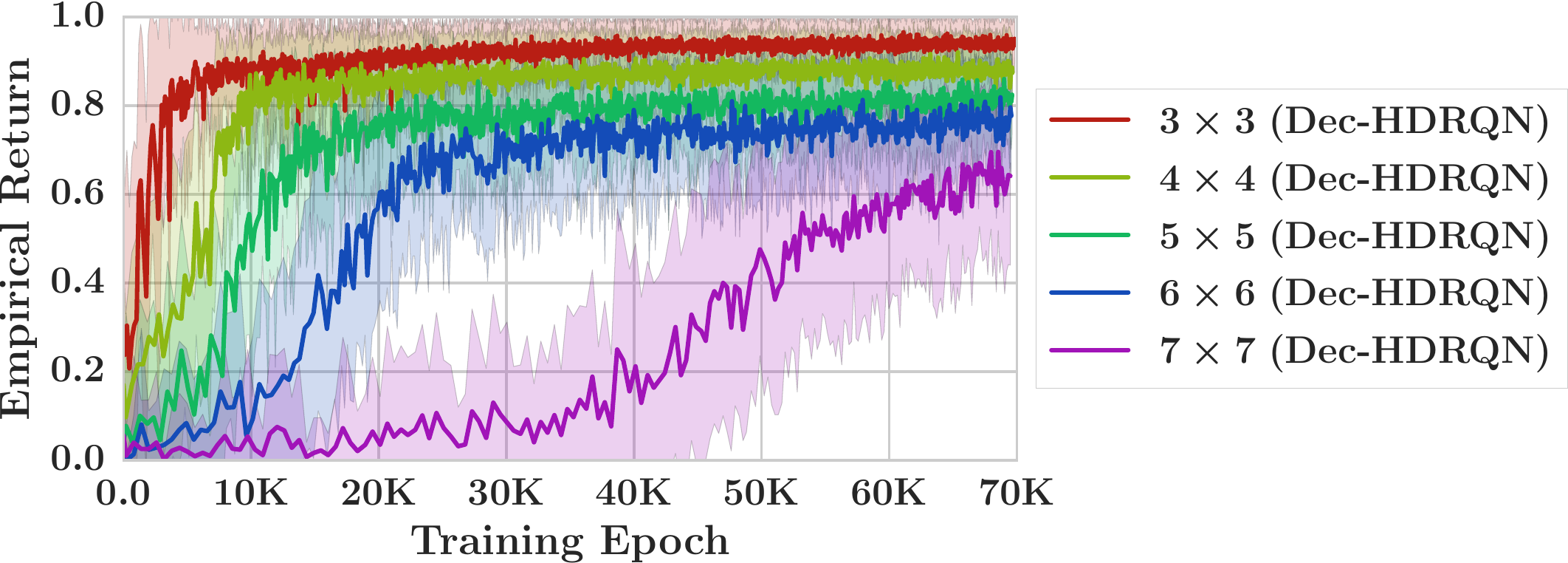}
		\caption{Dec-HDRQN (our approach) empirical returns during training.}
		\label{fig:2agt_mamt_hdrqn_v_appendix}
	\end{subfigure}
	\hfill
	\begin{subfigure}[t]{0.48\textwidth}
		\centering
		\includegraphics[width=1\linewidth]{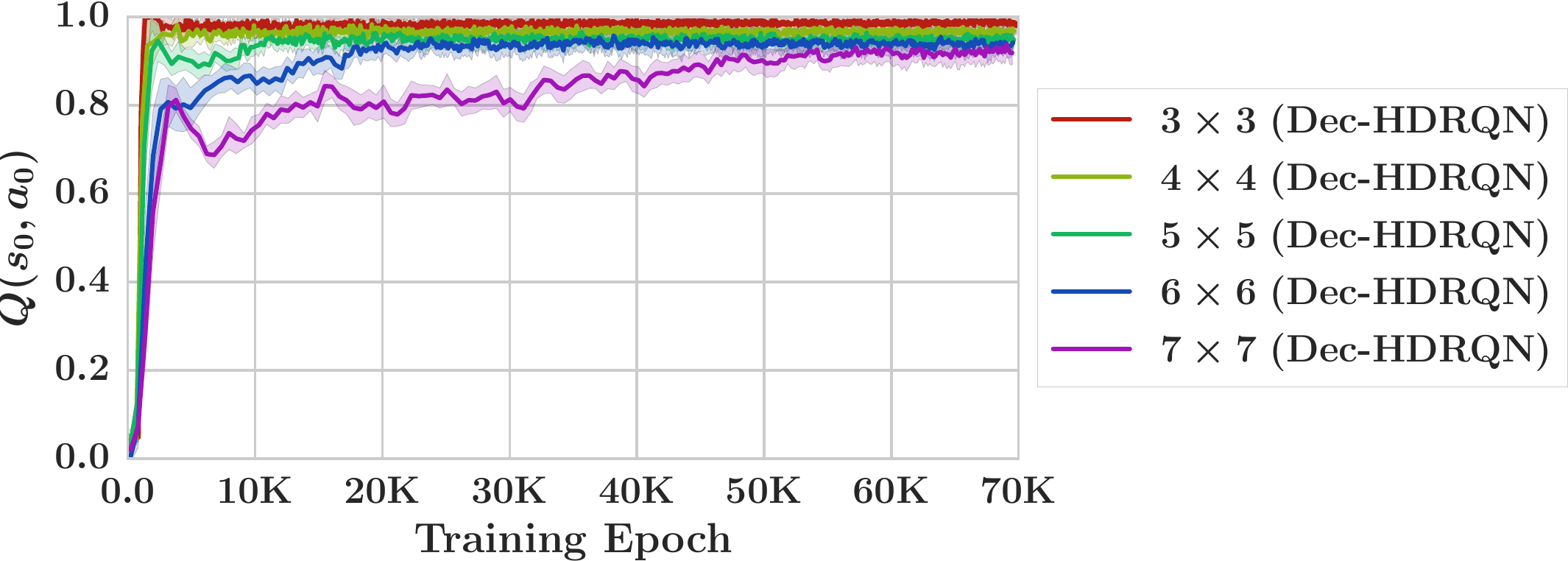}
		\caption{Dec-HDRQN (our approach) anticipated values during training.}
		\label{fig:2agt_mamt_hdrqn_q_appendix}
	\end{subfigure}
	\caption{MAMT domain results for Dec-DRQN and Dec-HDRQN, with 2 agents and $P_f = 0.3$. All plots conducted (at each training epoch) for a batch of 50 randomly-initialized episodes. Anticipated value plots (on right) were plotted for the exact starting states and actions undertaken for the episodes used in the plots on the left.}
	\label{fig:mamt_plots_comparisons_appendix}
\end{figure*}

\begin{figure*}
	\centering
	\includegraphics[width=0.7\linewidth]{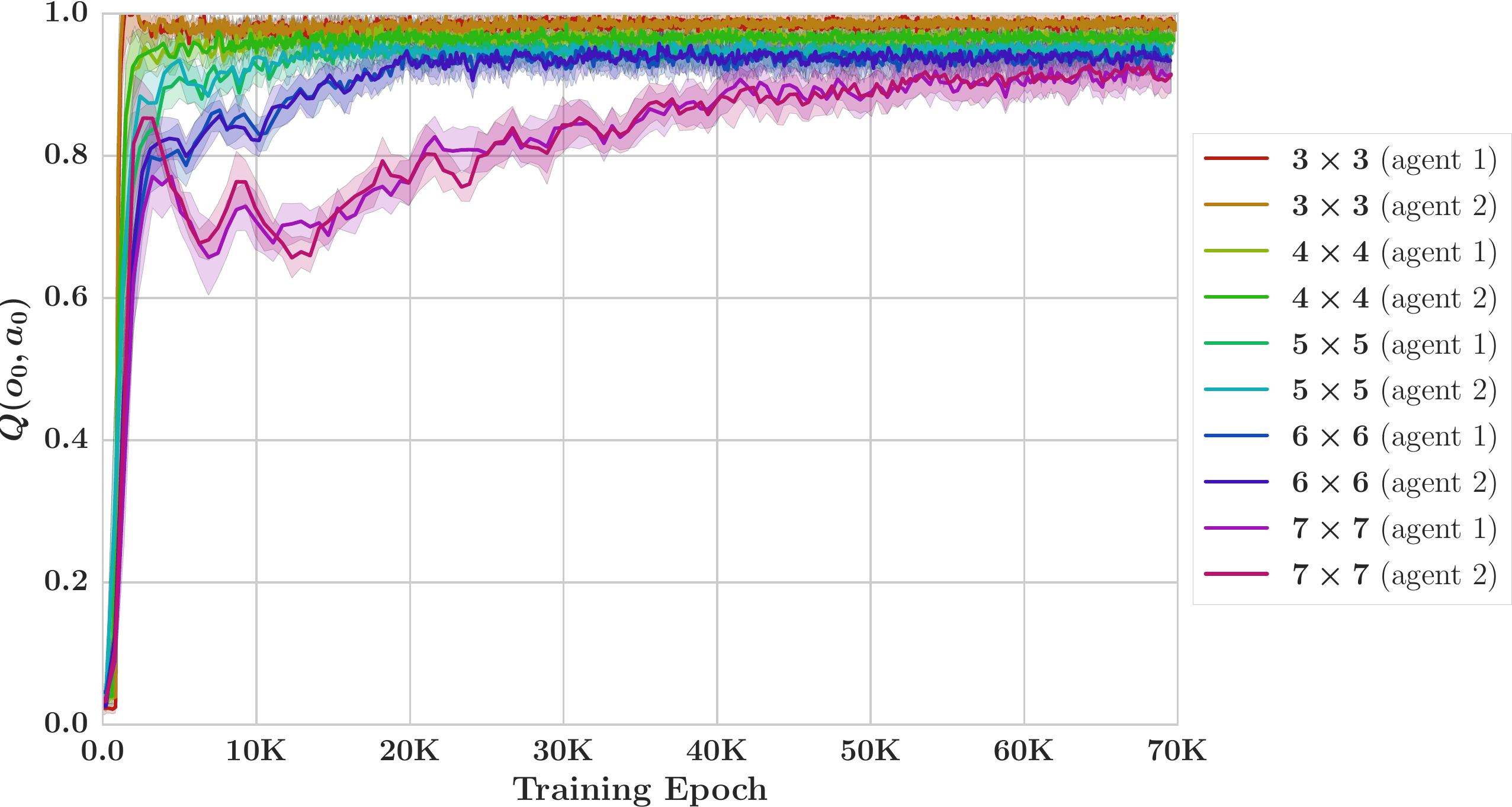}
\caption{Comparison of agents' anticipated value plots using Dec-HDRQN during training. MAMT domain, with 2 agents and $P_f = 0.3$. All plots conducted (at each training epoch) for a batch of 50 randomly-initialized episodes. For a given task, agents have similar anticipated value convergence trends due to shared reward; differences are primarily caused by random initial states and independently sampled target occlusion events for each agent. }
\label{fig:finalplot_multiagent_q_comparison}
\end{figure*}

\begin{figure*}[h]
	\centering
	\begin{subfigure}[h]{0.48\textwidth}
		\centering
		\includegraphics[width=1\linewidth]{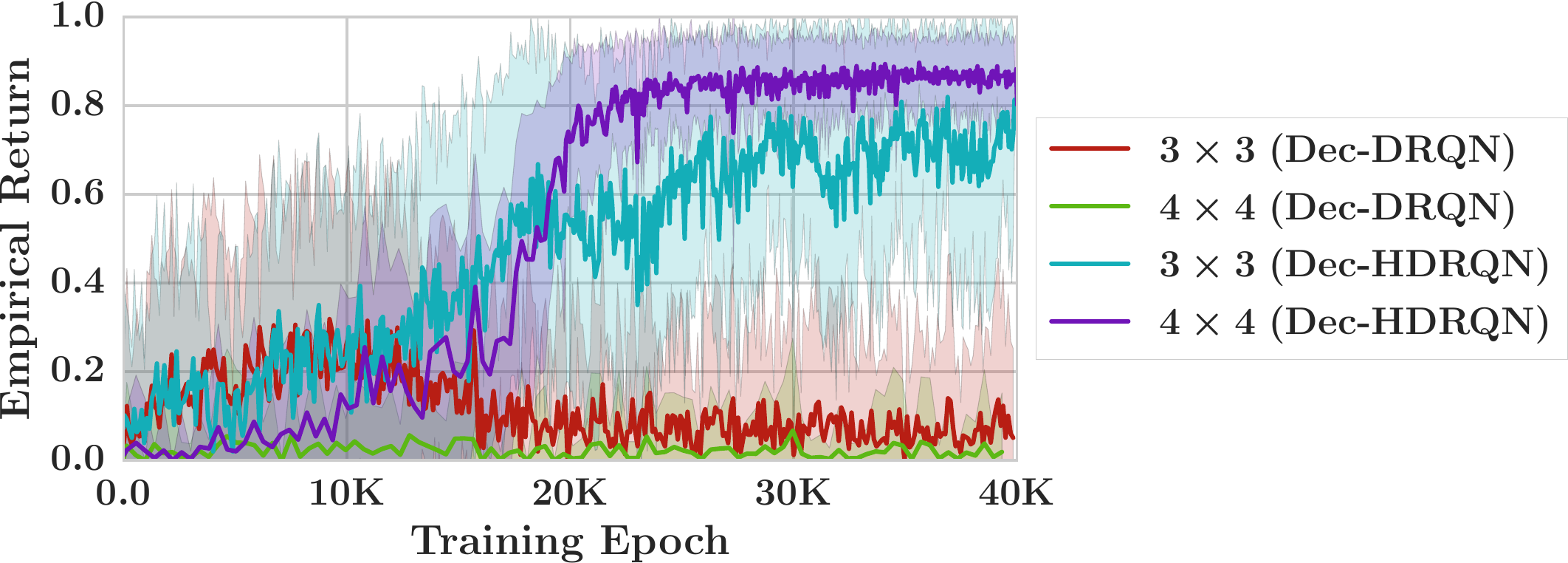}
		\caption{Empirical returns during training. For batch of 50 randomly-initialized games.}
		\label{fig:3agt_v_appendix}
	\end{subfigure}
	\hfill
	\begin{subfigure}[h]{0.48\textwidth}
		\centering
		\includegraphics[width=1\linewidth]{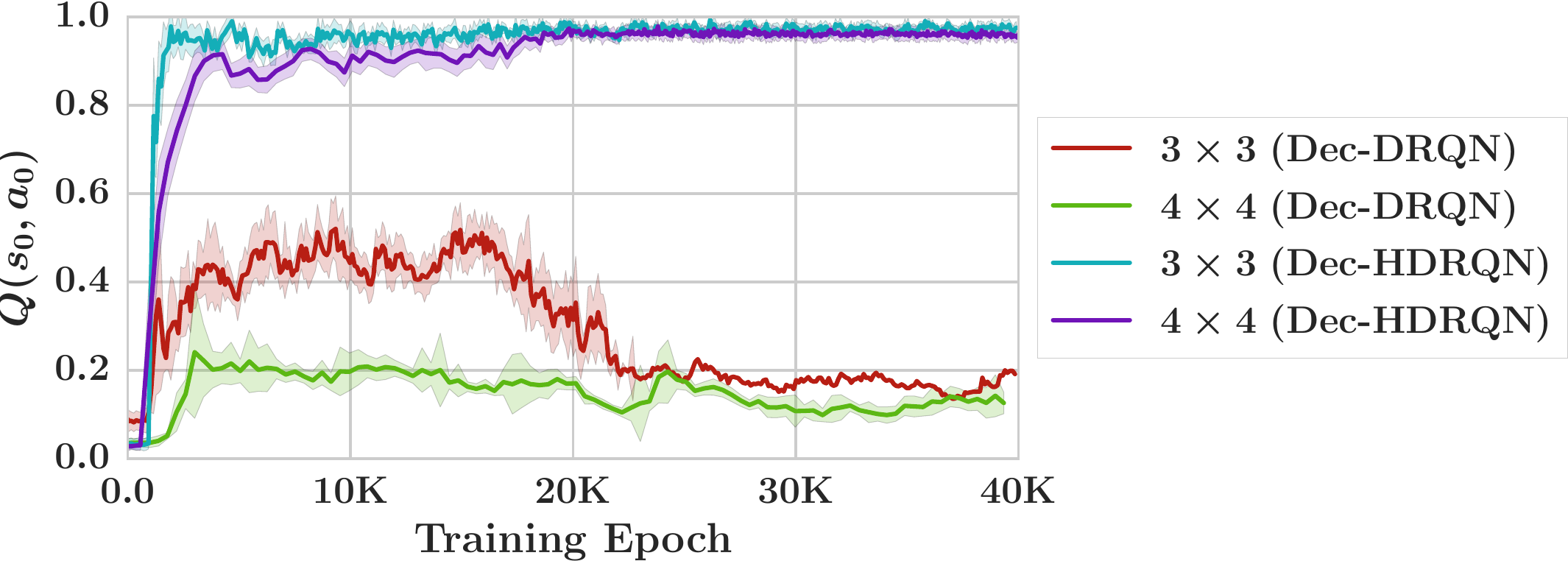}
		\caption{Anticipated values during training. For specific starting states and actions undertaken in the same 50 randomly-initialized games as \cref{fig:3agt_v_appendix}.}
		\label{fig:3agt_q_init_appendix}
	\end{subfigure}
	\caption{MAMT domain results for Dec-DRQN and Dec-HDRQN, with $n=3$ agents. $P_f = 0.6$ for the $3 \times 3$ task, and $P_f = 0.1$ for the $4 \times 4$ task.}
	\label{fig:mamt_plots_3agt_comparisons_appendix}
\end{figure*}

\twocolumn[] 
\section{Empirical Results: Learning Sensitivity to Dec-HDRQN Negative Learning Rate $\beta$ }

\begin{figure}[h]
	\centering
	\begin{subfigure}[h]{0.5\textwidth}
		\centering
		\includegraphics[width=1\linewidth]{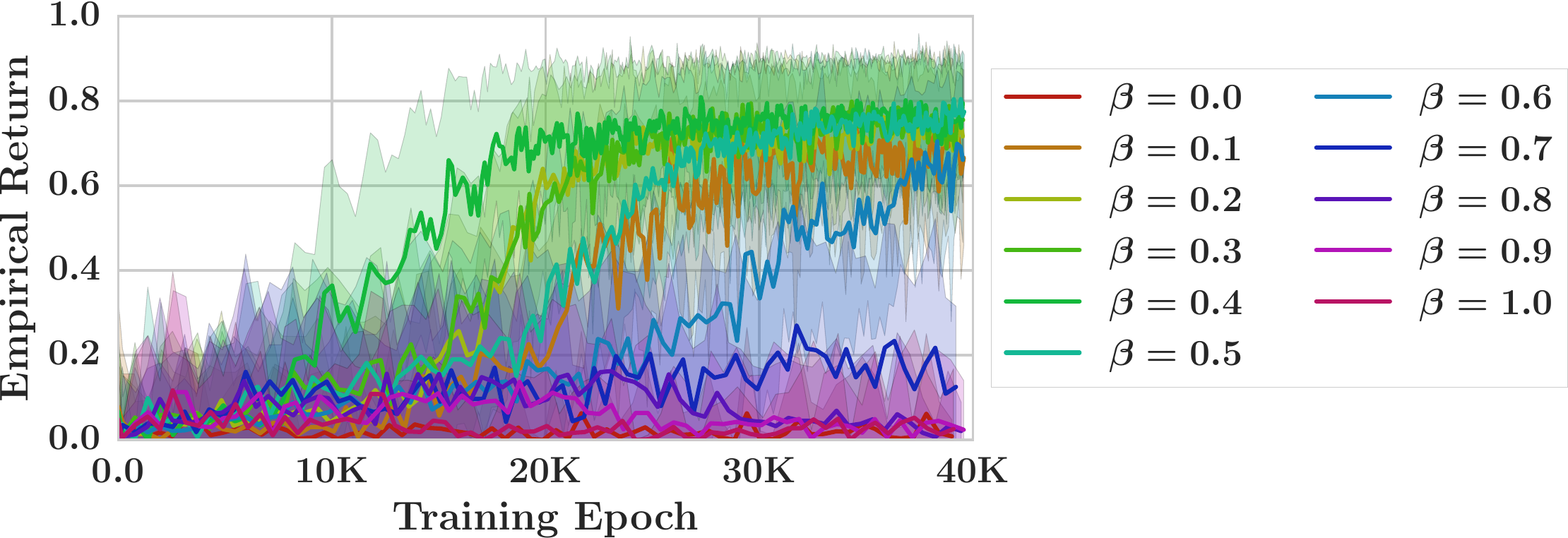}
		\caption{Sensitivity of Dec-HDRQN empirical returns to $\beta$ during training. For batch of 50 randomly-initialized games.}
		\label{fig:v_varying_beta}
	\end{subfigure}
	\hfill
	\begin{subfigure}[h]{0.5\textwidth}
		\centering
		\includegraphics[width=1\linewidth]{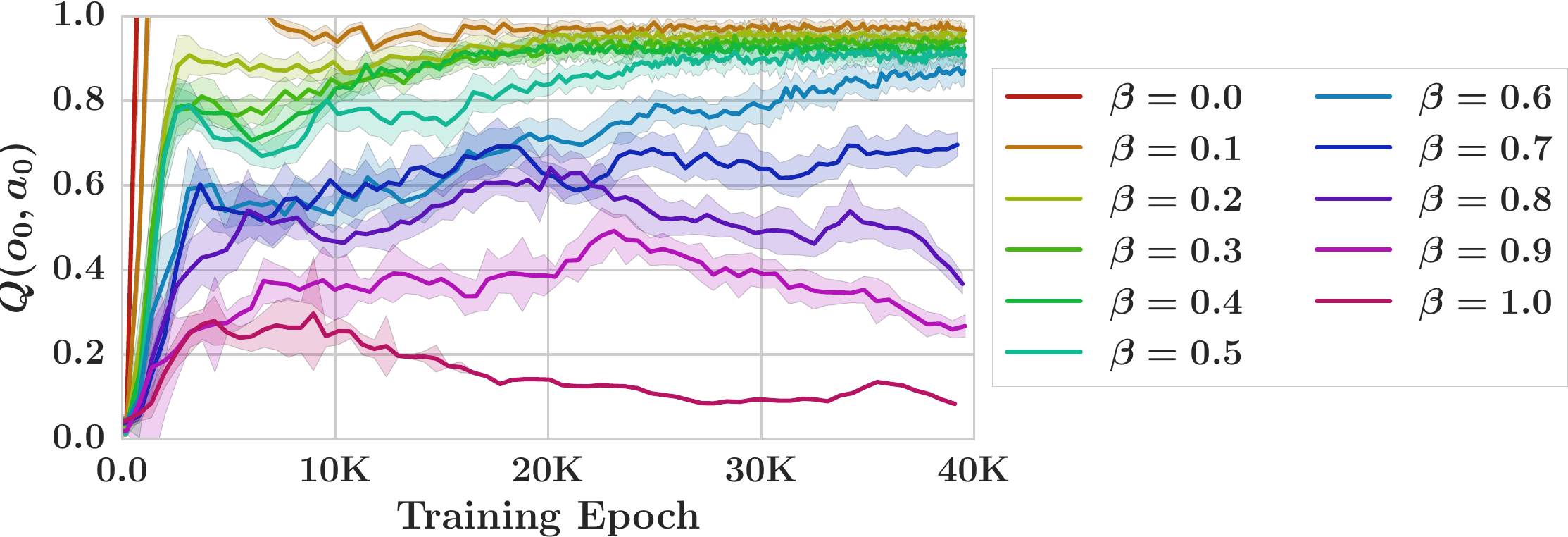}
		\caption{Sensitivity of Dec-HDRQN predicted action-values to $\beta$ during training. For specific starting states and actions undertaken in the same 50 randomly-initialized games of \cref{fig:v_varying_beta}.}
		\label{fig:q_init_varying_beta}
	\end{subfigure}
	\begin{subfigure}[h]{0.5\textwidth}
		\centering
		\includegraphics[width=1\linewidth]{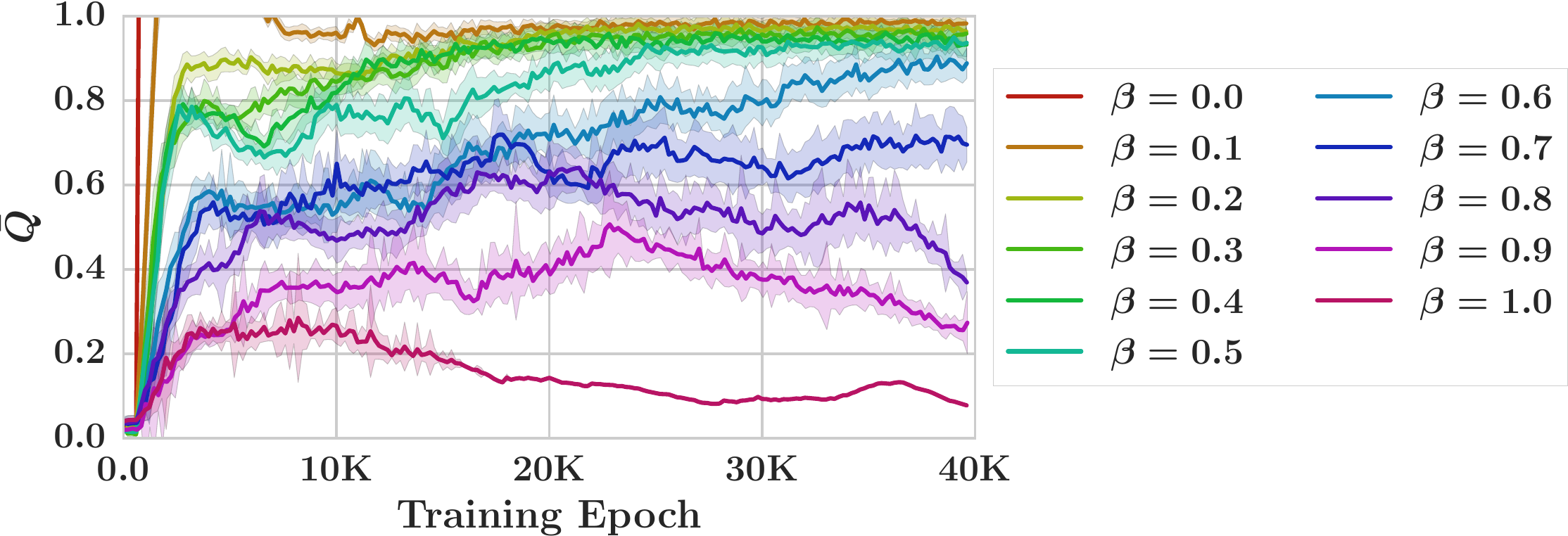}
		\caption{Sensitivity of Dec-HDRQN average Q values to $\beta$ during training. For random minibatch of 32 experienced observation inputs.}
		\label{fig:q_varying_beta}
	\end{subfigure}
	\caption{Learning sensitivity to $\beta$ for $6 \times 6$, 2 agent MAMT domain with $P_f=0.25$. All plots for agent $i=0$. $\beta=1$ corresponds to Decentralized Q-learning, $\beta=0$ corresponds to Distributed Q-learning (not including the distributed policy update step).}
	\label{fig:beta_sensitivity_values}
\end{figure}

\newpage

\section{Empirical Results: Learning Sensitivity to Dec-HDRQN Recurrent Training Tracelength Parameter $\tau$}
\begin{figure}[h]
	\centering
	\begin{subfigure}[h]{0.5\textwidth}
		\centering
		\includegraphics[width=1\linewidth]{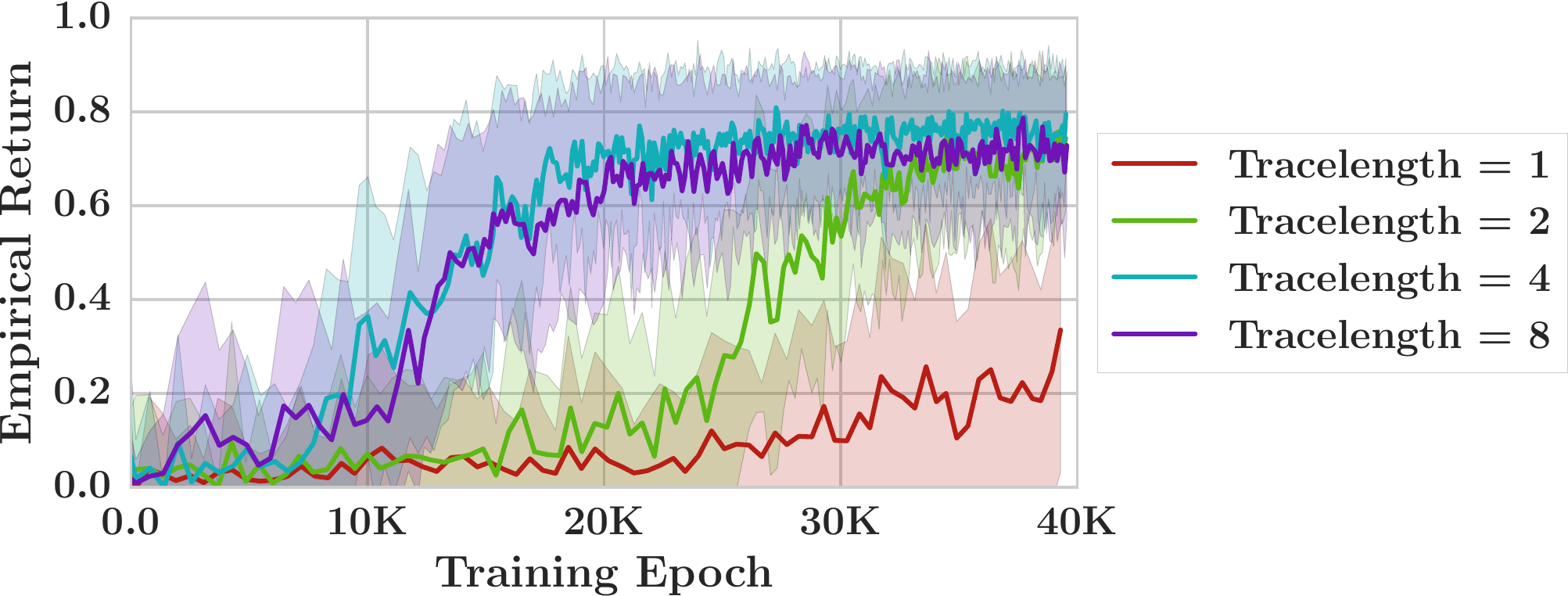}
		\caption{Dec-HDRQN sensitivity to tracelength $\tau$. 6x6 MAMT domain with $P_f=0.25$. }
		\label{fig:v_varying_tracelength_appendix}
	\end{subfigure}
	\begin{subfigure}[h]{0.5\textwidth}
		\centering
		\includegraphics[width=1\linewidth]{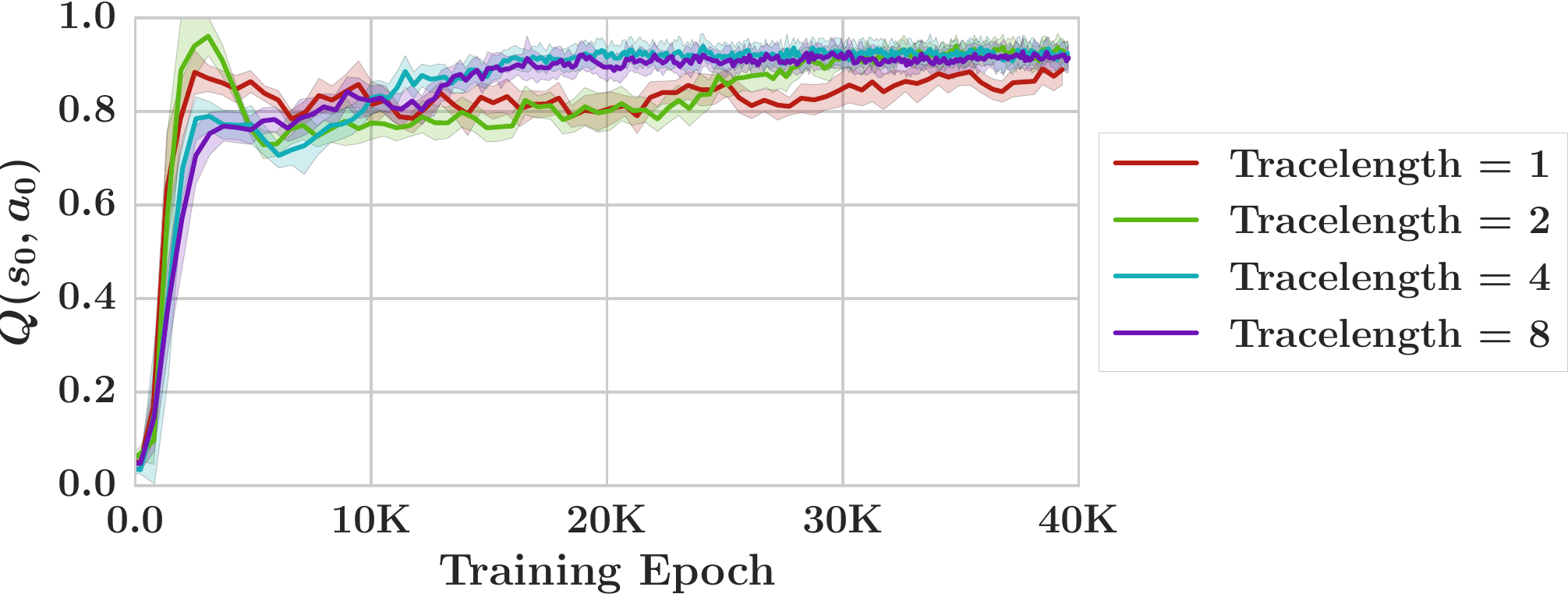}
		\caption{Sensitivity of Dec-HDRQN predicted action-values to recurrent training tracelength parameter. For specific starting states and actions undertaken in the same 50 randomly-initialized games of \cref{fig:v_varying_tracelength_appendix}.}
		\label{fig:q_init_varying_tracelength}
	\end{subfigure}
	\caption{Learning sensitivity to  recurrent training tracelength parameter for $6 \times 6$, 2 agent MAMT domain with $P_f=0.25$. All plots for agent $i=0$. $\beta=1$ corresponds to Decentralized Q-learning, $\beta=0$ corresponds to Distributed Q-learning (not including the distributed policy update step).}
	\label{fig:tracelength_sensitivity_values}
\end{figure}

\newpage
\section{Empirical Results: Multi-tasking Performance Comparison}
The below plots show multi-tasking performance of both the distillation and Multi-HDRQN approaches. Both approaches were trained on the $3 \times 3$ through $6 \times 6$ MAMT tasks. Multi-DRQN failed to achieve specialized-level performance on all tasks, despite 500K training epochs. By contrast, the proposed MT-MARL distillation approach achieves nominal performance after 100K epochs.

\begin{figure}[h]
	\centering
	\begin{subfigure}[t]{0.5\textwidth}
		\centering
		\includegraphics[width=1\linewidth]{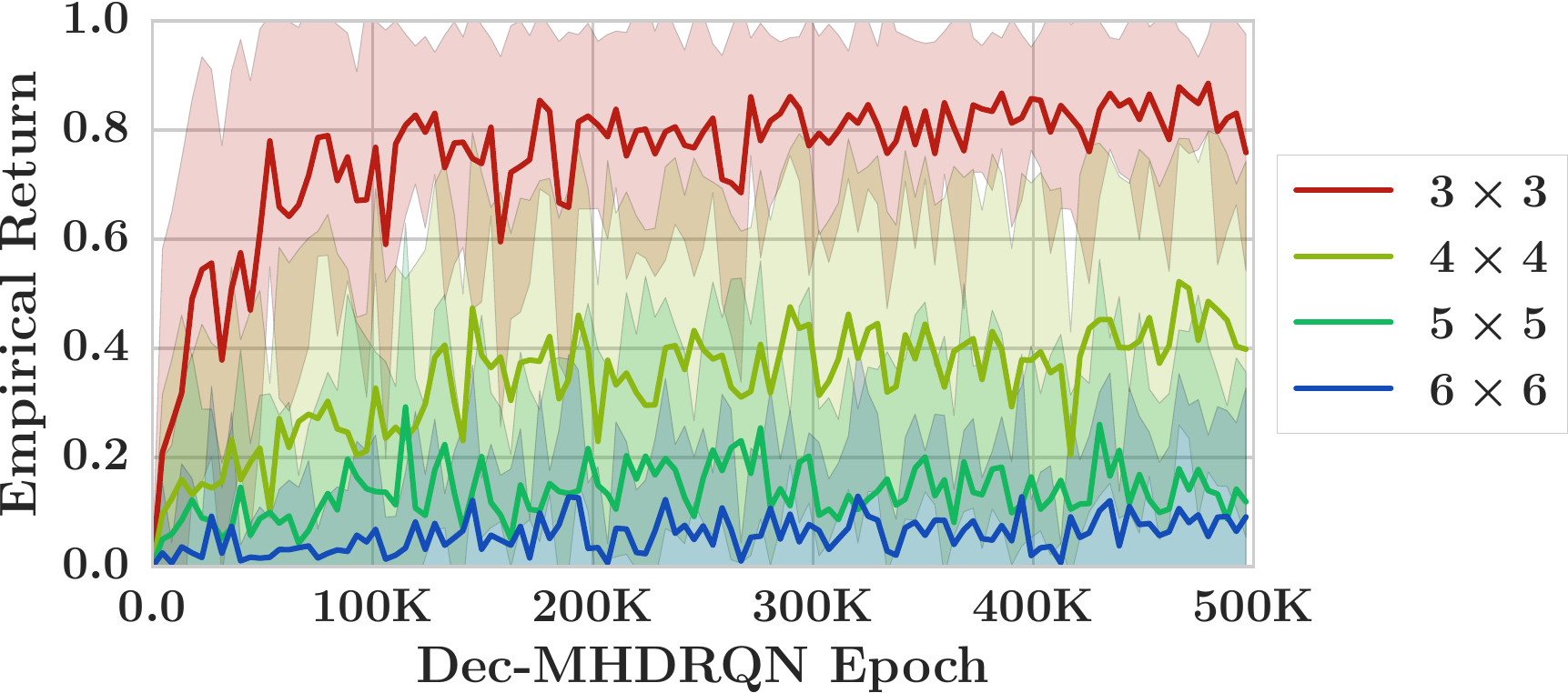}
		\caption{MT-MARL via Multi-HDQRN.}
		\label{fig:mhdrqn_appendix}
	\end{subfigure}	\\
	\begin{subfigure}[t]{0.5\textwidth}
		\centering
		\includegraphics[width=1\linewidth]{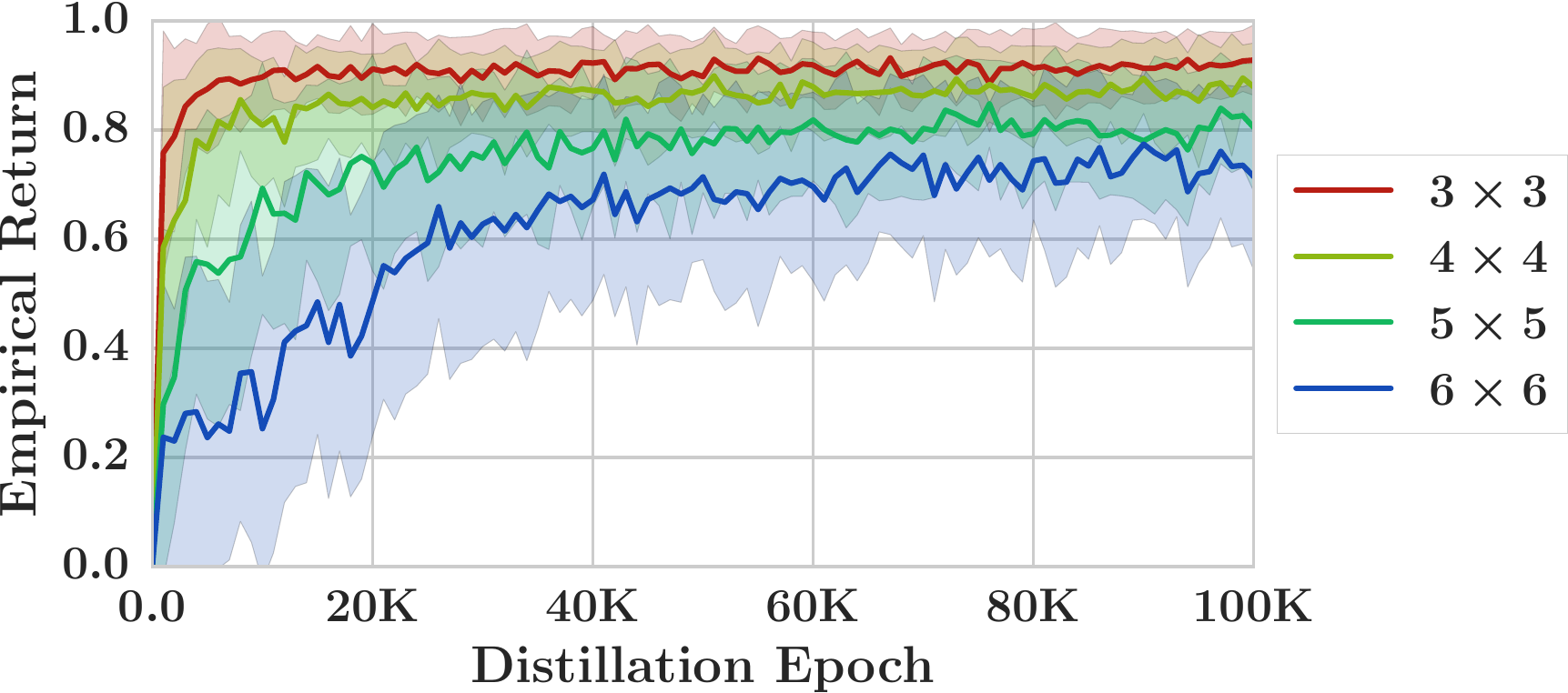}
		\caption{MT-MARL via specialized and distilled Dec-HDRQN.}
		\label{fig:distillation_appendix}
	\end{subfigure}
	\caption{Multi-task performance on MAMT domain, $n=2$ agents and $P_f=0.3$.}
	\label{fig:multitask_comparisons_appendix}
\end{figure}

\end{document}